%% file: main.tex
\documentclass[oneside, 11pt]{article}
\usepackage{iclr2024_conference,times}

\input{math_commands.tex}

\usepackage{textpos}
\usepackage{comment}
\usepackage{soul}
\usepackage[colorinlistoftodos]{todonotes}
\usepackage{booktabs}

\usepackage{tocloft} 
\cftpagenumbersoff{chapter}

\usepackage{tabularx,longtable,multirow}
\usepackage[margin=1cm]{caption}
\usepackage[margin=0.5cm]{subcaption}
\usepackage{fancyhdr} 
\usepackage{url} 
\usepackage[english]{babel}
\usepackage{amsmath, amssymb}
\usepackage{graphicx}
\usepackage{dsfont}
\usepackage{cancel} 
\usepackage{epstopdf} 
\usepackage{array}
\usepackage{latexsym}
\usepackage{xcolor}
\definecolor{light-gray}{gray}{0.2}
\definecolor{dark-gray}{gray}{0.15}
\usepackage{hyperref} 
\usepackage[shortlabels]{enumitem}

\usepackage{natbib}

\setcitestyle{authoryear,open={(},close={)}} 
\usepackage{bm}
\usepackage{amsfonts}

\usepackage{amsthm}
\theoremstyle{theorem}
\newtheorem{example}{Example}
\newtheorem{remark}{Remark}[section]

\newtheorem*{nonumbertheorem}{Theorem}

\usepackage[all]{hypcap}
\usepackage{pdflscape} 
\usepackage{afterpage}
\usepackage{adjustbox}

\usepackage{algorithm}
\usepackage{algpseudocode}
\usepackage{float}
\newfloat{algorithm}{t}{lop}

\usepackage{wrapfig}
\usepackage{lipsum}

\usepackage{bbm}

\newtheorem{lemma}{Lemma}
\newtheorem{theorem}{Theorem}

\def\sv{\mathbf{s}}

\def\R{\mathbb{R}}
\def\E{\mathbb{E}}
\def\T{\top}
\def\Ind{\mathbf{1}}
\def\eps{\varepsilon}

\newcommand{\EqRef}[1]{Eq.~\ref{#1}}

\title{\LARGE
Post-Hoc Bias Scoring is Optimal for Fair Classification
}
\author{Wenlong Chen \thanks{Equal contribution.}  \thanks{Work done during internship at ByteDance Research.} \\
Imperial College London\\
\texttt{wenlong.chen21@imperial.ac.uk} 
\And
Yegor Klochkov $^*$  \\
ByteDance Research \\
\texttt{yegor.klochkov@bytedance.com} 
\AND
Yang Liu \\
ByteDance Research \\
\texttt{yang.liu01@bytedance.com}
}

\iclrfinalcopy 
\begin{document}

\maketitle
\begin{abstract}
We consider a binary classification problem under group fairness constraints, which can be one of Demographic Parity (DP), Equalized Opportunity (EOp), or Equalized Odds (EO). We propose an explicit characterization of Bayes optimal classifier under the fairness constraints, which turns out to be a simple modification rule of the unconstrained classifier. Namely, we introduce a novel instance-level measure of bias, which we call \emph{bias score}, and the modification rule is a simple linear rule on top of the finite amount of {bias scores}.
Based on this characterization, we develop a \emph{post-hoc} approach that allows us to adapt to fairness constraints while maintaining high accuracy. In the case of DP and EOp constraints, the modification rule is thresholding a single bias score, while in the case of EO constraints we are required to fit a linear modification rule with 2 parameters.
The method can also be applied for composite group-fairness criteria, such as ones involving several sensitive attributes.
%
%
We achieve competitive {or better} performance compared to both \emph{in-processing} and \emph{post-processing} methods across three datasets: Adult, COMPAS, and CelebA. Unlike most \emph{post-processing} methods, we do not require access to sensitive attributes during the inference time.

\end{abstract}

\section{Introduction}
\input{sections/intro}

\section{Bayes optimal fairness-constrained classifier}
\input{sections/optimal_flip_rule}

\section{Methodology}
\input{sections/method}

\section{Experiments}
\input{sections/experiments}

\section{Related literature}\label{sec:related}

We mention some recent work on Bayes optimal classifiers under approximate fairness constraints. \cite{menon2018cost} consider cost-aware binary classification, followed by a series of results that characterize sensitive attribute-aware fair classifiers. These include DP constraints with either binary or multi-class target and either binary or multi-class sensitive attribute  \cite{denis2021fairness, xian2023fair, gaucher2023fair}.  \cite{zeng2022bayes} additionally considers other group fairness metrics. These papers do not cover the cases of Equalized Odds and Composite Criterion.
Implementations in \cite{denis2021fairness, zeng2022bayes} are based on closed form expressions of the threshold weights, 
while our MBS treats the bias scores as features. 

\section{Conclusion}
\input{sections/conclusion}


\section*{Acknowledgments}

We are grateful to the anonymous referees for valuable feedback that helped to improve the presentation, extend experiments and theoretical results. We are thankful to our former collegue Muhammad Faaiz Taufiq for plenty of fruitful conversations. We also thank Yingzhen Li for her valuable comments on the manuscript.

\section*{Reproducibility statement}
Details for the experimental set-up are provided in the beginning of Section~\ref{sec:experiment}, and the code can be found at \url{https://github.com/chenw20/BiasScore}.

\bibliography{main}

\newpage
\appendix

\section{Proof of Theorem~\ref{main_thm}}
\input{appendix/proof}

\section{Practical Algorithms}
\input{appendix/algorithm}


\section{Ablation Studies}\label{sec:}
\input{appendix/ablation}

\section{Results in Tables}
\input{appendix/Tables}

\newpage

\section{Sensitivity Analysis for Composite Criterion}
\input{appendix/sensitivity_analysis}

\newpage
\section{Comparison with Reduction Method \citep{agarwal2018reductions}}
\input{appendix/reduction}

\end{document}

%% file: math_commands.tex
\usepackage{amsmath,amsfonts,bm}

\def\eqref#1{equation~\ref{#1}}

\def\1{\bm{1}}

\def\eps{{\epsilon}}

\DeclareMathAlphabet{\mathsfit}{\encodingdefault}{\sfdefault}{m}{sl}
\SetMathAlphabet{\mathsfit}{bold}{\encodingdefault}{\sfdefault}{bx}{n}

\newcommand{\E}{\mathbb{E}}

\newcommand{\R}{\mathbb{R}}

\DeclareMathOperator{\sign}{sign}

%% file: sections/intro.tex
With ML algorithms being deployed in more and more decision-making applications, it is crucial to ensure fairness in their predictions. Although the debate on what is fairness and how to measure it is ongoing \citep{caton2023fairness}, 
group fairness measures are utilized in practice due to the simplicity of their verification \citep{chouldechova2017fair, hardt2016equality}, and they conform to the intuition that predictions should not be biased toward a specific group of the population. In practice, it is desirable to train classifiers satisfying these group fairness constraints while maintaining high accuracy.

Training classifiers that maintain competitive accuracy and satisfy group fairness constraints remains a challenging problem, and it often requires intervention during the training time. A popular {approach} \citep{zafar2017fairness, zafar2019fairness} suggests relaxing these constraints of a discrete nature to score-based differentiable constraints, thus {utilizing}
gradient-based optimization methods. {This} {approach is} very flexible and can be used in a broad set of applications \citep{donini2018empirical, cotter2019optimization, rezaei2021robust, wang2021fair, zhu2023weak}. Another popular method suggests dynamically reweighting observations during training \citep{agarwal2018reductions}.
In vision tasks, researchers propose to use more sophisticated techniques, such as synthetic image generation \citep{ramaswamy2021fair} and contrastive learning \citep{park202fair}.

Another strategy proposes to modify unconstrained classifiers in a \emph{post-hoc} manner \citep{hardt2016equality, jiang2019wasserstein, jang2022group}. Unlike the \emph{in-processing} methods, these methods allow one to adapt to fairness constraints after the model is trained. These modifications are much cheaper and more feasible in industrial settings, where very large datasets are utilized and complicated algorithms are used to train the target classifier. However, the existing solutions typically require knowledge of the sensitive attribute during the inference time. For instance, 
\citet{hardt2016equality} propose to modify a score-based classifier in the form $ \hat{Y}(X) = \Ind\{ R(X) > t\} $ to a group-specific thresholding rule $ \check{Y}(X, A) = \Ind\{ R(X) > t_{A} \} $. Similar approaches are also taken by \citet{jiang2019wasserstein, jang2022group}. This is impractical for real-world applications where sensitive attributes during inference are inaccessible due to privacy protection.

Most of the existing methods aim at debiasing a classifier, whether with an \emph{in-processing} or \emph{post-processing} method. We ask a more general question: how can we flexibly adjust a classifier to achieve the best accuracy for a given level of fairness? For a binary classification problem, \citet{menon2018cost} study this question from a theoretical perspective, that is, when one knows the ground truth distribution $p(Y, A |X)$, they derive the Bayes-optimal classifier satisfying fairness constraints.
Unfortunately, \citet{menon2018cost} only cover two cases of fairness measures: Demographic Parity and Equalized Opportunity. In this paper, we close this gap and derive the Bayes-optimal classifier for general group fairness metrics, which include the case of Equalized Odds. 
Our analysis also allows using composite fairness criteria that involve more than one sensitive attribute at the same time.  See other related references in Section~\ref{sec:related}. 

We interpret our solution as a modification of (unconstrained) Bayes optimal classifier based on a few values that we term \emph{``bias scores''}, which in turn can be thought of as a measure of bias on instance level. 
For instance, think of reducing the gender gap in university admissions. \citet{bhattacharya2017university} show that such gap reduction typically happens at the expense of applicants with borderline academic abilities. In terms of classification (passed/ not passed), this corresponds to the group where we are least certain in the evaluation of one's academic abilities. This suggests that evaluation of bias on instance level should not only take into account prediction and group membership, but also uncertainty in the prediction of target value. Our \emph{bias score} not only conforms to this logic, but thanks to being part of Bayes optimal classifier, it is also theoretically principled. In particular, for the case of Demographic Parity constraints, we show that the optimal constrained classifier can be obtained by modifying the output of the unconstrained classifier on instances with largest bias score. When Equalized Odds constraints are imposed, or more generally a composite criterion, the optimal modification is a linear rule with two or more bias scores.

Based on our characterization of the optimal classifier, we develop a practical procedure to adapt any score-based classifier to fairness constraints. 
In Section~\ref{sec:experiment}, we show various experiments across three benchmarks: Adults, COMPAS, and CelebA. We are able to achieve better performance than the in-processing methods, despite {only} being able to adapt to the group-fairness constraints {after training}. We also provide competitive results when compared to post-processing methods \citep{hardt2016equality, jiang2019wasserstein}, which require knowledge of sensitive attribute during inference.

We summarize our contributions as follows.
    Firstly, we characterize the Bayes optimal classifier under group fairness constraints, 
    which generalizes \citet{menon2018cost} in the sense that \citet{menon2018cost} can be viewed as a special case in our framework, where the constraint is only a single fairness criterion (e.g. Demographic Parity). Nevertheless, our formulation is more convenient and intuitive thanks to the interpretable \emph{bias score}. 
    Secondly, our characterization can work with composite fairness criterion (e.g. Equalized Odds) as constraint, which has not been established before to our knowledge. Thirdly, based on this characterization, we propose a post-processing method that can flexibly adjust the trade-off between accuracy and fairness and does not require access to test sensitive attributes. Empirically, our method achieves competitive or better performance compared with baselines.



%

\subsection{Preliminaries}

In this work, we consider binary classification, which consists of many practical applications that motivate machine fairness research \citep{caton2023fairness}.  We want to construct a classifier $\hat{Y} = \hat{Y}(X)$ for a target variable $ Y \in \{0, 1\} $ based on the input $X$. Apart from the accuracy of a classifier, we are concerned with fairness measurement, given that there is a sensitive attribute $A$, with some underlying population distribution over the triplets $(X, Y, A) \sim \Pr$ in mind. We assume that the sensitive attribute is binary as well. We generally focus on three popular group-fairness criteria:
\begin{itemize}[leftmargin=15pt, topsep=-1pt]
    \item \textbf{Demographic Parity (DP)} \citep{chouldechova2017fair} is concerned with equalizing the probability of a positive {classifier} output in each sensitive group,
\end{itemize}
    \begin{equation}\label{dp_def}
        DP(\hat{Y}; A) = \left| \Pr(\hat{Y} = 1 | \; A = 0) - \Pr(\hat{Y} = 1 | A = 1)\right| .
    \end{equation}
\begin{itemize}[leftmargin=*]    
    \item \textbf{Equalized Odds (EO)} was introduced in \citet{hardt2016equality}. Unlike DP which suffers from explicit trade-off between fairness and accuracy in the case where $Y$ and $A$ are correlated, this criterion is concerned with equalizing the false positive and true positive rates in each sensitive group,
    \begin{equation}\label{eo_def}
        EO(\hat{Y}; A) = \max_{y =0, 1} \left| \Pr(\hat{Y} = 1 | A = 0, Y = y) - \Pr(\hat{Y} = 1 | A = 1, Y = y)\right|.
    \end{equation}
    \item \textbf{Equality of Opportunity (EOp)} was also introduced by \citet{hardt2016equality}, and it measures the disparity only between true positive rates in the two sensitive groups,
    \begin{equation}\label{eop_def}
        EOp(\hat{Y}; A) = \left| \Pr(\hat{Y} = 1 | A = 0, Y = 1) - \Pr(\hat{Y} = 1 | A = 1, Y = 1)\right|.
    \end{equation}
\end{itemize}

%% file: sections/optimal_flip_rule.tex
\label{sec: optimal_flip_rule}
For a given distribution over $\Pr \sim (X, Y)$, the {Bayes} optimal unconstrained classifier has the form $ \hat{Y}(X) = \Ind\{ p(Y = 1 |X) > 0.5\} $ in the sense that it achieves maximal accuracy.
Although it is not generally possible to know these ground-truth conditional probabilities ($p(Y = 1 |X)$) in practice, such characterization allows one to train probabilistic classifiers, typically with cross-entropy loss. Here we ask what is the optimal classifier when the group fairness restrictions are imposed. Namely, which classifier $\check{Y}(X)$ maximizes the accuracy $ Acc(\check{Y}) = \Pr(\check{Y} = Y) $ under the restriction that a particular group fairness measure is below a given level $\delta > 0$.

We are interested in either DP, EOp, or EO constraints, as described in the previous section. For the sake of generality, we consider the following case of \emph{composite criterion}. Suppose, we have several sensitive attributes $A_1, \dots, A_K$ and for each of them we fix values $a_k, b_k$, so that we need to equalize the groups $\{A_k = a_k\}$ and $\{A_k = b_k\}$. In other words, our goal is to minimize a \emph{composite criterion} represented by a  maximum over a set of disparities,
\begin{equation}\label{composite criteria}
    CC(\check{Y}) = \max_{j = 1, \dots, K} \left| \Pr(\check{Y} = 1 | A_k = a_k) -  \Pr(\check{Y} = 1 | A_k = b_k)  \right|,
\end{equation}
This general case covers DP, EOp, and EO, as well as composite criteria involving more than one sensitive attribute. Let us give a few examples.

\begin{example}[Demographic Parity]
For the case of DP (see \EqRef{dp_def}), it is straightforward: take $A_1 = A \in \{0, 1\}$, $a_1 = 0, b_1 = 1$, and  then with $K = 1$, $CC(\check{Y}) = DP(\check{Y})$.
\end{example}

\begin{example}[Equalized Opportunity and Equalized Odds]
Suppose we have a sensitive attribute $A \in \{0, 1\}$, then the Equalized Opportunity criterion (\EqRef{eo_def}), can be written in the form of \EqRef{composite criteria} with $A_1 = (A, Y)$, $a_1 = (0, 1)$, and $b_1 = (1, 1)$.

For the Equalized Odds, we can write it as a composite criterion with $K = 2$ by setting $A_1 = A_2 = (A, Y)$, and setting $a_1 = (0, 0)$, $b_1 = (1, 0)$, $a_2 = (0, 1)$, $b_2 = (1, 1)$ in \EqRef{composite criteria}.
\end{example}

\begin{example}[Two and more sensitive attributes]
We could be concerned with fairness with respect to two sensitive attributes $A, B$ simultaneously (for instance, gender and race). In this case, we want to minimize the maximum of two Demographic Parities, which looks as follows
\[
    \max\{ \left| \Pr(\hat{Y} = 1 | A = 0) -  \Pr(\hat{Y} = 1 | A = 1)  \right|, \left| \Pr(\hat{Y} = 1 | B = 0) -  \Pr(\hat{Y} = 1 | B = 1)  \right| \}.
\]
If we have three DPs, we will have $K=3$ in \EqRef{composite criteria}; if we are interested in a maximum over EO's for two different sensitive attributes, we would have $K = 4$, etc.
\end{example}

Given a specified fairness level $\delta > 0$, we want to find the optimal classifier $\check{Y}(X)$, possibly randomized, that is optimal under the composite criterion constraints
\begin{equation}\label{constrained_problem}
  \max\;\;  Acc(\check{Y}) = \Pr(\check{Y} = Y)  \qquad \text{s.t.} \qquad CC(\check{Y}) \leq \delta .
\end{equation}
We will be searching the solution in the form of modification of the {Bayes} optimal unconstrained classifier. Recall that in our notation, $\hat{Y} = \hat{Y}(X)$ denotes the {Bayes} optimal unconstrained classifier $\Ind\{ p(1 |X) > 0.5\}$. We ``reparametrize'' the problem by setting $\kappa(X) = \Pr(\check{Y} \neq \hat{Y} |X)$ as the target function. In other words, given an arbitrary function $\kappa(X) \in [0, 1]$, we can define the modification $\check{Y}(X)$ of $\hat{Y}(X)$ by drawing $Z \sim Be(\kappa(X))$ and outputting
\[
\check{Y} = \left\{
\begin{aligned}
\hat{Y}, \qquad &Z = 0 \\
1 - \hat{Y}, \qquad &Z = 1
\end{aligned}
\right.
\]
We call such function $\kappa(X)$ a \emph{modification rule}. 
With such reparameterization, the accuracy of a modified classifier can be rewritten as
\[
    Acc(\check{Y}) = Acc(\hat{Y}) - \int \eta(X) \kappa(X) d\Pr(X),
\]
where $ \eta(X) =2p(Y=\hat{Y}|X)-1$ represents the confidence of the Bayes optimal unconstrained classifier $\hat{Y}$ on the instance $X$ (see Section~\ref{simple_proof_for_dp_case} for detailed derivation), A similar representation holds for the value of the composite criterion. Specifically, recall the criterion is of the form $ CC(\check{Y}) = \max_{k\leq K}|C_{k}(\check{Y})| $, where
\[
    C_k(\check{Y}) = \Pr(\check{Y} = 1 | A_k = a_k) -  \Pr(\check{Y} = 1 | A_k = b_k) \, .
\]
We can rewrite it as
\begin{align}
     C_k(\check{Y}) &\phantom{:}= C_k(\hat{Y}) - \int f_k(X) \kappa(X) d\Pr(X), \label{c_j_integral}\\
     f_k(X) &:= (2 \hat{Y} - 1) \left[ \frac{p(A_k = a_k |X)}{\Pr(A_k = a_k)} - \frac{p(A_k = b_k |X)}{\Pr(A_k = b_k)} \right] . \label{f_j_definition}
\end{align}
These two expressions suggest that modifying the answer on a point with low confidence $\eta(X)$ makes the least losses in the accuracy, while modifying the answers with higher absolute value of $f_k(X)$ makes largest effect on the parity values, although this has to depend on the sign. This motivates us to define modification rule on the relative score,
\begin{equation}\label{bias_score}
   \text{(Instance-Level Bias Score):} \qquad s_{k}(X) = \frac{f_k(X)}{\eta(X)},
\end{equation}
which we refer to as \emph{bias score}. It turns out that the optimal modification rule, i.e. one corresponding to the constrained optimal classifier in \EqRef{constrained_problem}, is a simple linear rule with respect to the given $K$ bias scores. We show this rigorously in the following theorem. We postpone the proof to the appendix, Section~\ref{simple_proof_for_dp_case}.

\begin{theorem}\label{main_thm}
Suppose that all functions $f_k, \eta $ are square-integrable and the scores $ s_{k}(X) = f_{k}(X) / \eta(X) $ have joint continuous distribution. Then, for any $\delta > 0$, there is an optimal solution defined in \EqRef{constrained_problem} that is obtained with a modification rule of the form,
\begin{equation}\label{optimal_kappa_from_theorem}
    \kappa(X) = \mathbf{1}\left\{ \sum_{k} z_{k} s_{k}(X) > 1 \right\} .
\end{equation}
\end{theorem}

This result suggests that for the case of DP, EOp, or EO, given the {ground-truth} probabilities $p(Y, A|X)$, we only need to fit $1$ parameter for either of Demographic Parity and Equalized Opportunity, which essentially corresponds to finding a threshold, and fit a linear rule in two dimensions for the Equalized Odds. 
Below we consider each of the three fairness measures in detail.

\paragraph{Demographic Parity.}
In the case of DP constraint, we have a single bias score of the form,
\begin{equation}
    s(X) = \frac{1}{\eta(X)} (2 \hat{Y} - 1) \left[ \frac{p(A = 0 |X)}{\Pr(A = 0)} - \frac{p(A = 1 |X)}{\Pr(A = 1)} \right],
\end{equation}
and since there is only one score, the modification rule is a simple threshold rule of this bias score $ \kappa(X) = \Ind\{ s(X) / t > 1 \} $. We note that $t$ can be positive or negative, depending on which group has the advantage, see Section~\ref{simple_proof_for_dp_case}. This allows one to make linear comparison of fairness on the instance level. That is, departing from a fairness measure defined on a group level, we derive a bias score that measures fairness on each separate instance.
%
For example, in the context of university admissions \citep{bhattacharya2017university}, our bias score conforms with the following logic: it is more fair to admit a student who has high academic performance (lower $\eta(X)$) than one who has borderline performance  (higher $\eta(X)$) even though they are both equally likely to come from the advantageous group (same $f(X)$). 
We note that the problem of measuring fairness and bias on instance level has recently started to attract attention, see \citet{wang2022understanding, yao2023understanding}.

\paragraph{Equality of Opportunity.} Here we also have the advantage of having a simple threshold rule, corresponding to the score function,
\[
    s(X) = \frac{1}{\eta(X)} (2 \hat{Y} - 1) \left[ \frac{p(A = 0, Y = 1 | X)}{\Pr(A = 0, Y = 1)} -\frac{p(A = 1, Y = 1 | X)}{\Pr(A = 1, Y = 1)} \right] \,.
\]

\begin{remark}[Comparison to group-aware thresholding]
Let us consider the case where there is a one-to-one correspondence $A = A(X)$. This is equivalent to the case of observed sensitive attribute, and it is trivial to check that our method turns into a group-aware thresholding and becomes oblivious \citep{hardt2016equality, jang2022group} (i.e., the flipping rule doesn't depend on interpretation of individual features $X$ any more). Indeed, in such case we have $ p(A = a, Y = 1 | X) = p(Y = 1 |X) \Ind\{A(X) = a\} $, therefore $ s(X) = \frac{p(Y = 1 | X) }{2 p(Y = 1|X) - 1} \left[ a_0 \Ind\{A(X) = 0\} + a_1 \Ind\{ A(X) = 1\}   \right] $, where $ a_0 = 1/ \Pr(A = 0, Y = 1) > 0 $ and $a_1 = - 1/\Pr(A = 1, Y = 1) < 0$. Then for a given $t_{\delta} $, the final decision rule turns into $ \check{Y}(X, A) = \Ind\{ p(Y = 1| X) > t_{A}\} $, where $ t_A = 0.5 + a_{A} / (4t_{\delta} - 2a_{A})$.
\end{remark}

\paragraph{Equalized Odds.} Let us consider the case of optimizing under Equalized Odds constraint in detail. In this case, we need to know the {ground-truth} conditional probabilities $ p(Y, A |X)$, and we obtain two scores for $k = 0, 1$,
\begin{equation}\label{s_score_eo}
    s_{k}(X) = \frac{1}{\eta(X)} \{2 \hat{Y} - 1\} \left[ \frac{p(Y = k, A = 0 | X)}{\Pr(Y = k, A = 0)} - \frac{p(Y = k, A = 1 | X)}{\Pr(Y = k, A = 1)} \right]
\end{equation}
Our goal is then to find a linear rule in the bias embedding space $ (s_0(X), s_1(X)) $, which on validation, achieves the targeted equalized odds, while maximizing the accuracy. Notice that here the problem is no longer a simple threshold choice as in the case of DP-constrained classifier. We still need to fit a fairness-constrained classifier, only we have dramatically reduced the complexity of the problem to dimension $K = 2$, and we only have to fit a linear classifier.

We {demonstrate the modification rule in the case of EO constraints with} the following synthetic data borrowed from \citet{zafar2019fairness} [Section~{5.1.2}] {as example}: 
\begin{equation}\label{zafar_synthetic}
\begin{aligned}
    p(X| Y = 1, A = 0) &= \mathcal{N}([2, 0], [5,1; 1, 5]), &
    p(X| Y = 1, A = 1) &= \mathcal{N}([2, 3], [5,1; 1, 5]), \\
    p(X| Y = 0, A = 0) &= \mathcal{N}([-1, -3], [5,1; 1, 5]), &
    p(X| Y = 0, A = 1) &= \mathcal{N}([-1, 0], [5,1; 1, 5]). 
\end{aligned} 
\end{equation}
We sample $500$, $100$, $100$, $500$ points from each of groups $(Y, A) = (1, 0)$, $(1,1)$, $(0,0)$, $(0,1)$, respectively, so that $Y$ and $A$ are correlated. Next, we fit a logistic linear regression with $4$ classes {to estimate} $p(Y, A | X) $ and calculate the scores according to the formulas \EqRef{s_score_eo}. In Figure~\ref{fig:syn_scores}, we show the scatter plot of the scores $(s_0(X), s_1(X))$, with the corresponding group marked by different colors. Figures~\ref{fig:syn_delta_0_15}-\ref{fig:syn_delta_0_01} show the optimal flipping rule, with color encoding $\kappa(X)$ evaluated with the discretized version of the linear program, while the red line approximately shows the optimal linear separation plane. We observe that some of the points that we had to flip for the restriction $ EO \leq \delta = 0.15 $, are unflipped back when the restriction is tightened to $ EO \leq \delta = 0.01 $. It indicates that unlike in the case of DP restriction, there is no unique score measure that can quantify how fair is the decision made by a trained classifier.



%% file: sections/method.tex
\label{sec: method}
In practice, especially for deep learning models, unconstrained classifiers are usually of the form $ \hat{Y} = \Ind\{ \hat{p}(Y | X) > 0.5\}$, with the conditional probability trained using the cross-entropy loss. Our characterization of the optimal modification rule naturally suggests a practical post-processing algorithm that takes fairness restriction into account{:} assume that we are given an \emph{auxiliary} model for either $ \hat{p}(A|X)$ (in the case of DP constraints) or $ \hat{p}(Y, A|X)$ (in the case of EOp and EO constraints). We then treat these estimated conditionals as ground-truth conditional distributions, plugging them into \EqRef{bias_score} to compute the bias scores, and modify the prediction $\hat{Y}$ correspondingly with a linear rule over these bias scores. We propose to fit the linear modification rule using a labeled validation set. We call this approach  \emph{Modification with Bias Scores} (MBS) and it does not require knowing test set sensitive attribute since the bias scores are computed based on the estimated conditional distributions related to sensitive attribute ($ \hat{p}(A|X)$ or $ \hat{p}(Y, A|X)$) instead of the empirical observations of sensitive attribute.  
Here we demonstrate the algorithms in detail for the two cases where DP and EO are the fairness criteria.

\paragraph{Post-processing algorithm with DP constraints.}
In this case, we assume that we have two models $\hat{p}(Y|X)$ and $\hat{p}(A|X)$ (which in the experiments are fitted over the training set, but can be provided by a third party as well) to estimate the ground truth $p(Y|X)$ and $p(A|X)$ respectively. We then define the bias score as follows:
\begin{equation}
        \hat{s}(X)=\frac{\hat{f}(X)}{\hat{\eta}(X)} = \frac{\{2 \hat{Y}(X) - 1\} \left[ \frac{\hat{p}(A = 0 | X)}{\widehat{\Pr}(A = 0)} -\frac{\hat{p}(A = 1 | X)}{\widehat{\Pr}(A = 1)} \right]}{2 \hat{p}(Y=\hat{Y}(X) |X) - 1},
\end{equation}
where $\widehat{\Pr}(A = i)$ ($i=0, 1$) can be estimated by computing the ratio of the corresponding group in the training set. We search for the modification rule of the form $ \kappa(X) = \Ind\{ \hat{s}(X) / t > 1 \}$, so that the resulting $\check{Y}_t(X) = \Ind\{ \hat{s}(X) / t \leq 1 \} \hat{Y}(X) + \Ind\{ \hat{s}(X) / t > 1 \} (1 - \hat{Y}(X)) $ satisfies the DP constraint, while maximizing the accuracy. For this, we assume that we are provided with a labeled validation dataset $ \{(X_i, Y_i, A_i)\}_{i = 1}^{N_{val}}$ and we choose the threshold $t$ such that the validation accuracy is maximized, while the empirical DP evaluated on it is $\leq \delta$. To find the best threshold value, we simply need to go through all $N_{val}$ candidates $ t = \hat{s}(X_i) $, which can be done in $O(N_{val}\log N_{val})$ time. See detailed description in Algorithm~\ref{alg: dp} in the appendix, Section~\ref{appendix: alg}.

\paragraph{Post-processing algorithm with EO constraints.}
In this case, we require an auxiliary model $\hat{p}(Y,A|X)$ with four classes. This allows us to obtain the 2D estimated bias score $(\hat{s}_0(X), \hat{s}_1(X))$, where
\begin{equation}
    \begin{aligned} \hat{s}_0(X)&=\frac{\hat{f}_0(X)}{\hat{\eta}(X)} = \frac{\{2 \hat{Y}(X) - 1\} \left[ \frac{\hat{p}(A = 0, Y = 0 | X)}{\widehat{\Pr}(A = 0, Y = 0)} -\frac{\hat{p}(A = 1, Y = 0 | X)}{\widehat{\Pr}(A = 1, Y = 0)}\right]}{2 \hat{p}(Y=\hat{Y}(X) |X) - 1} ,\\
        \hat{s}_1(X_i)&=\frac{\hat{f}_1(X)}{\hat{\eta}(X_i)} = \frac{\{2 \hat{Y}(X) - 1\} \left[ \frac{\hat{p}(A = 0, Y = 1 | X)}{\widehat{\Pr}(A = 0, Y = 1)} -\frac{\hat{p}(A = 1, Y = 1 | X)}{\widehat{\Pr}(A = 1, Y = 1)}\right]}{2 \hat{p}(Y=\hat{Y}(X) |X) - 1} ,
    \end{aligned}
\end{equation}
where each of the $\widehat{Pr}(A = a, Y = y)$ is again estimated from training set. We are searching for a linear modification rule $ \kappa(X) = \Ind\{ a_0 \hat{s}_0(X) + a_1 \hat{s}_1(X) > 1\}$ that for a given validation set satisfies the empirical EO constraint while maximizing the validation accuracy. We consider two strategies to choose such a linear rule.

In the first approach, we take a subsample of points $\{(\hat{s}_0(X_m'), \hat{s}_1(X_m'))\}_{m=1}^M$ of size $M \leq N_{val}$ and consider all $M (M- 1) / 2$ possible linear rules passing through any two of these points. For each of these rules, we evaluate the EO and accuracy on validation set, then choose maximal accuracy among ones satisfying $EO \leq \delta$. The total complexity of this procedure is $O(M^2 N_{val})$. A formal algorithm is summarized in Algorithm \ref{alg: eo} in the appendix, Section~\ref{appendix: alg}.

\begin{figure}[t]
    \centering
    \begin{subfigure}{0.25\textwidth}
        \includegraphics[width=\textwidth]{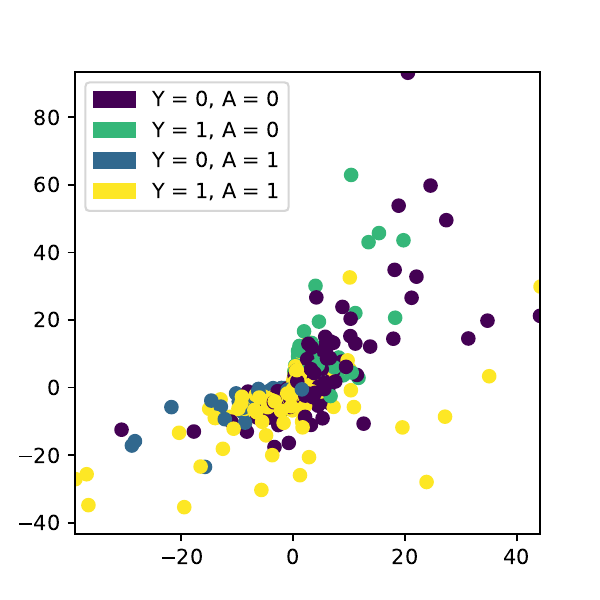}
        \caption{}
        \label{fig:syn_scores}
    \end{subfigure}
    \begin{subfigure}{0.25\textwidth}
        \includegraphics[width=\textwidth]{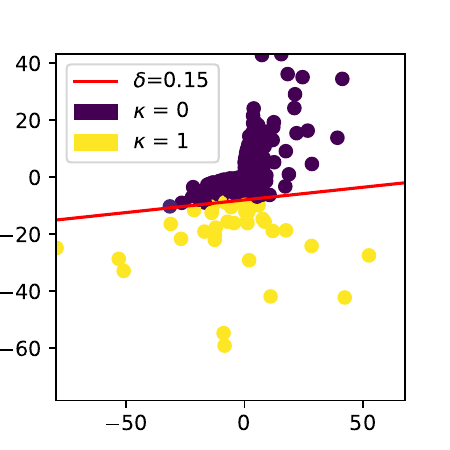}
        \caption{}
        \label{fig:syn_delta_0_15}
    \end{subfigure}
    \begin{subfigure}{0.25\textwidth}
        \includegraphics[width=\textwidth]{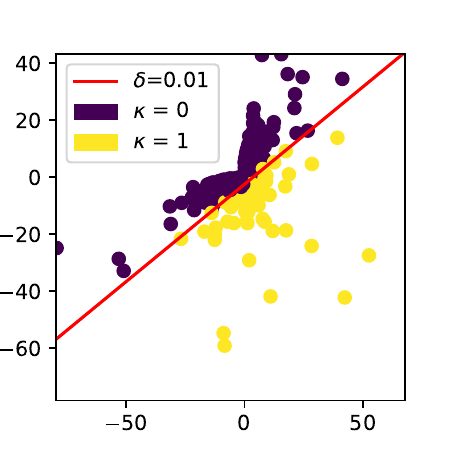}
        \caption{}
        \label{fig:syn_delta_0_01}
    \end{subfigure}
    \caption{(a) Scatter plot of the scores for synthetic distribution~\EqRef{zafar_synthetic}. (b) Separation plane for the optimal flipping rule $\kappa$ corresponding to $ EO \leq \delta = 0.15 $ and (c) $\delta = 0.01$.}
    \label{fig:zafar_example}
\end{figure}

We also consider a simplified version, where we fix a set of $K$ equiangular directions $ w = (\cos(2\pi j/K), \sin(2\pi j/K)) $ for $j = 0, \dots, K-1$. Then, for a score $ w_0 \hat{s}_0(X) + w_1 \hat{s}_1(X)  $ we simply need to choose a threshold, following the procedure in the DP case, where we evaluate the EO and accuracy dynamically. The time complexity in $O(K N_{val}\log N_{val})$, see details in Algorithm~\ref{alg: eo2}, Section~\ref{appendix: alg} in the appendix.

\begin{remark}
    Note that as functions of $X$, the probabilities ${p}(Y|X)$ and $p(Y, A|X)$ must agree in order for Theorem~\ref{main_thm} to hold, in the sense that $ p(Y|X) = p(Y, 0|X) + p(Y, 1|X)$. {However, the form of the algorithms itself, which only requires ``plug-in'' estimators $\hat{p}(Y|X)$ and $\hat{p}(Y,A|X)$ for ground-truth $p(Y|X)$ and $p(Y, A|X)$ respectively, does not require strict agreement between $\hat{p}(Y|X)$ and $\hat{p}(Y,A|X)$ for it to run. In practice, we can employ $\hat{p}(Y|X)$, $\hat{p}(Y, A|X)$ that were trained separately, and we can run the algorithm even in the case where the auxiliary model $\hat{p}(Y, A|X)$ was pretrained by a third party on another dataset similar to the one of interest.} E.g., we conduct an experiment where the auxiliary model is based on CLIP \citep{radford2021learning} in Section~\ref{sec:clip}.
\end{remark}


\paragraph{Sensitivity analysis.}
Here we investigate the sensitivity of our method to inaccuracy in the {estimated} conditional distributions. We first provide theoretical analysis, which takes into account two sources of error: approximation of the {ground-truth} conditional distributions $p(Y|X)$ and $p(A|X)$ by, say parametric models, and the sampling error in evaluation of the accuracy and DP on validation. {The details regarding the conditions and omitted constants are deferred to the appendix. See precise formulation and the proof in Section~\ref{appendix: sensitivity_analysis}, which also includes general composite criterion.}

\begin{nonumbertheorem}[Informal]
Suppose that we have estimations of conditional distributions $ \hat{p}(Y|X) $, $\hat{p}(A|X)$ {(or $\hat{p}(Y, A|X)$ when using EO constraint)}, and assume that
\begin{align*}
    \E |\hat{p}(Y|X) - p(Y |X)| \leq \eps,
    \qquad
    \E |\hat{p}(A|X) - p(A |X)| &\leq \eps \;\;&\text{(for DP),}\\
    {\E |\hat{p}(Y,A|X) - p(Y,A |X)|} &{\leq \eps}\;\;&{\text{(for EO).}}
\end{align*}
Let $ \check{Y} $ be the algorithm obtained with Algorithm~\ref{alg: dp} {(Algorithm~\ref{alg: eo} for EO)}. Then, with high probability
\begin{align*}
    Acc(\hat{Y}) - Acc(\check{Y})  \lesssim \sqrt{\eps} + \sqrt{(\log N_{val}) / N_{val}},
    \qquad
    &DP(\check{Y}) - \delta \lesssim \sqrt{(\log N_{val}) / N_{val}} \, , \\
    (&{EO(\check{Y}) - \delta \lesssim \sqrt{(\log N_{val}) / N_{val}}) \, .}
\end{align*}
\end{nonumbertheorem}


Moreover, we include three ablation studies in appendix \ref{appendix: ablation}, where less accurate $\hat{p}(Y|X)$, $\hat{p}(A|X)$ or $\hat{p}(Y, A|X)$ are deployed to examine the robustness of the post-processing modification algorithm. We find that our post-processing algorithm still retains the performance even when $\hat{p}(Y|X)$, $\hat{p}(A|X)$ or $\hat{p}(Y, A|X)$ are moderately inaccurate.

%% file: sections/experiments.tex
\label{sec:experiment}

We evaluate 
MBS on real-world binary classification tasks with the following experimental set-up.

\paragraph{Datasets.} We consider three benchmarks:
\begin{itemize}[leftmargin=15pt, topsep=-1pt]
\item \textbf{Adult Census} \citep{kohavi1996scaling}, a UCI tabular dataset where the task is to predict whether the annual income of an individual is above \$50,000. We randomly split the dataset into a training, validation and test set with 30000, 5000 and 10222 instances respectively. We pre-process the features according to \citet{lundburg2017unified} and the resulting input $X$ is a 108-dimensional vector. We use ``Gender'' as the sensitive attribute;
\item \textbf{COMPAS} \citep{angwin2015machine}, a tabular dataset where the task is to predict the recidivism of criminals. The dataset is randomly split into a training, validation and test set with 3166, 1056 and 1056 instances respectively. The input $X$ consists of 9 normal features (e.g. age and criminal history) and we choose ``Race" as the sensitive attribute;
\item \textbf{CelebA} \citep{liu2015deep}, a facial image dataset containing 200k instances each with 40
binary attribute annotations. We follow the experimental setting as in \citet{park202fair}: we choose ``Attractive", ``Big nose", and ``Bag Under Eyes" as target attributes, and choose ``Male" and ``Young" as sensitive attributes, yielding 6 tasks in total, and we use the original train-validation-test split.
\end{itemize}

\paragraph{Network architectures and hyperparameters.} We use an MLP for Adult Census and COMPAS datasets, with hidden dimension chosen to be 8 and 16 respectively. For each CelebA experiment, we use a ResNet-18 \citep{he2016deep}. For experiments with DP constraints, we train two models $\hat{p}(Y|X)$ and $\hat{p}(A|X)$ to predict the target and sensitive attributes respectively, while for experiments with EO constraints, we only train one model but with four classes $\hat{p}(Y, A|X)$, with each class corresponding to one element in the Cartesian product of target and sensitive attributes.

\defcitealias{zafar2017fairness}{Z17}
\defcitealias{jiang2019wasserstein}{J19}
\defcitealias{hardt2016equality}{H16}
\defcitealias{park202fair}{P22}

\paragraph{Baselines.} For experiments on Adult Census and COMPAS, we compare MBS with \citet{zafar2017fairness} (\citetalias{zafar2017fairness})
, \citet{jiang2019wasserstein} (\citetalias{jiang2019wasserstein}) (post-processing version, for experiments with DP constraints) and \citet{hardt2016equality} (\citetalias{hardt2016equality}) (for experiments with EO constraints). For CelebA, we additionally compare with \citet{park202fair} (\citetalias{park202fair}), which is a strong baseline tailored to fair facial attribute classification on CelebA. We report the averaged performance from 3 independent runs for all methods.

\paragraph{Evaluations \& metrics.} We consider both Demographic Parity (DP) and Equalized Odds (EO) as fairness criteria. We select the modification rules $\kappa(X)$ over the validation set according to the algorithms in Section \ref{sec: method}. We consider three levels of constraints for the fairness criteria: $\delta=10\%\text{, } 5\%\text{, } \text{ and } 1\%$, and we set $M$ in Algorithm~\ref{alg: eo} described in Section~\ref{sec: method} to be 3000, 600 and 5000 for experiments with EO as fairness criterion on Adult Census, COMPAS and CelebA\textcolor{blue}{,} respectively. Then we report the test set accuracy and DP/EO computed based on the post-processed test predictions after modification according to $\kappa(X)$. 

\begin{figure}[t]
    \centering
    \begin{subfigure}{0.245\textwidth}
        \includegraphics[width=\textwidth]{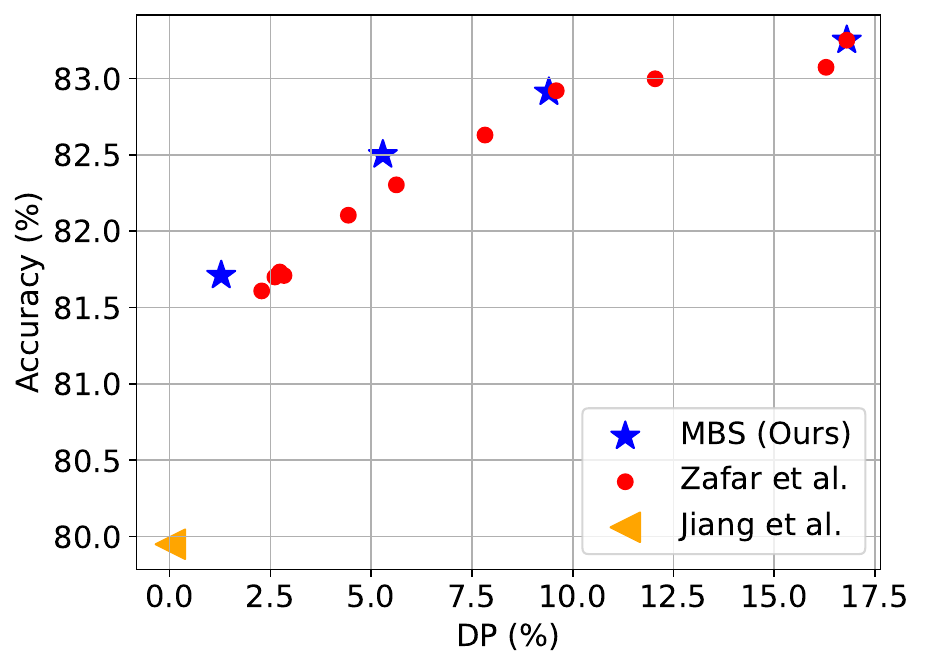}
        \caption{}\label{fig_dp_adult}
    \end{subfigure}
    \begin{subfigure}{0.245\textwidth}
        \includegraphics[width=\textwidth]{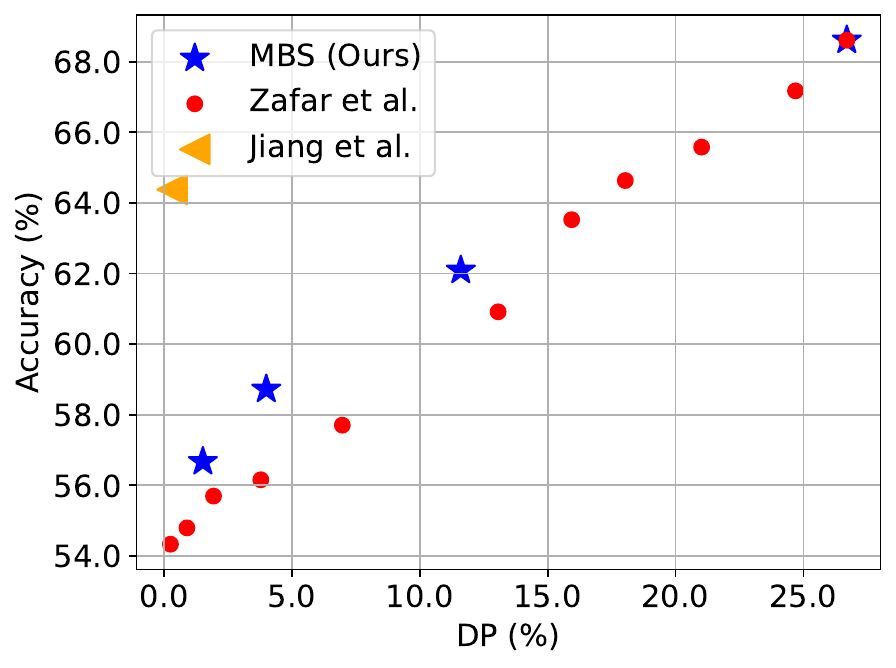}
        \caption{}\label{fig_dp_compas}
    \end{subfigure}
    \centering
    \begin{subfigure}{0.245\textwidth}
        \includegraphics[width=\textwidth]{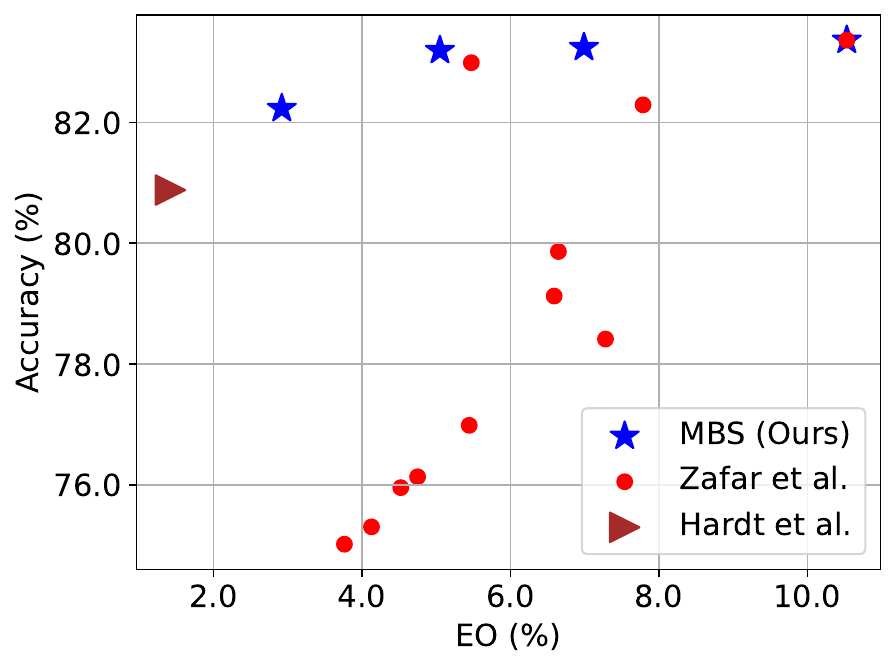}
        \caption{}\label{fig_eo_adult}
    \end{subfigure}
    \begin{subfigure}{0.245\textwidth}
        \includegraphics[width=\textwidth]{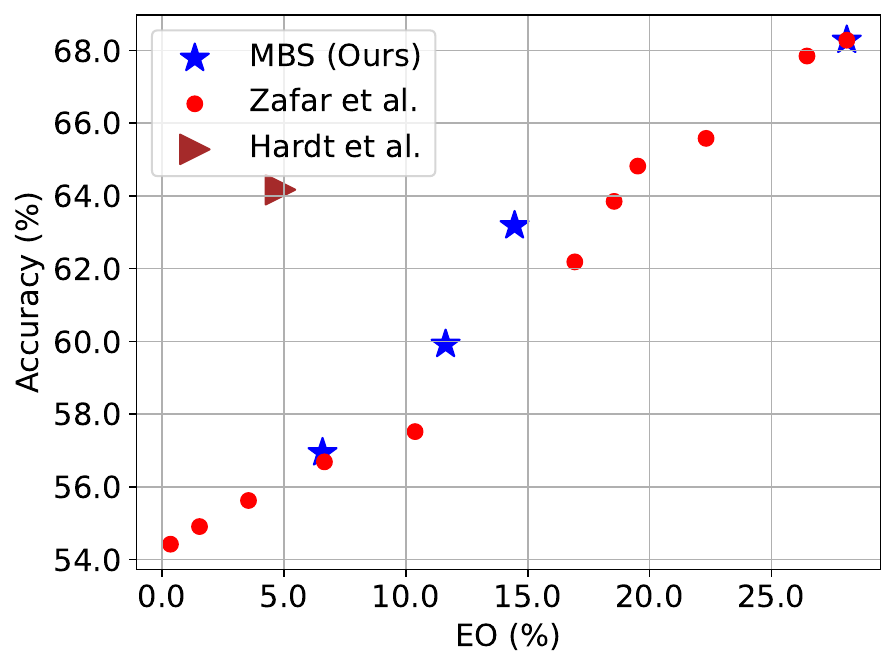}
        \caption{}\label{fig_eo_compas}
    \end{subfigure}
    \caption{Accuracy (\%) vs Demographic Parity (DP) (\%) trade-offs on (a) Adult Census and (b) COMPAS; Accuracy (\%) vs Equalized Odds (EO) (\%) trade-offs on (c) Adult Census and (d) COMPAS. Desired $\delta=\infty\text{ (unconstrained), }10\%\text{, } 5\%\text{, } \text{ and } 1\%$.}
    \label{fig: eo_dp_results}
\end{figure}

\subsection{Experiments with DP as fairness criterion}
We consider experiments with DP as fairness criterion on Adult Census and COMPAS datasets, and we compare MBS with \citetalias{zafar2017fairness} and \citetalias{jiang2019wasserstein}. The results are reported in Figures~\ref{fig_dp_adult}-\ref{fig_dp_compas}. One can see MBS consistently outperforms 
\citetalias{zafar2017fairness}
for both datasets in the sense that given different desired levels ($\delta$'s) of DP, MBS tends to achieve higher accuracy. Furthermore, while \citetalias{zafar2017fairness}
requires retraining the model each time to achieve a different trade-off between accuracy and DP, we are able to flexibly balance between accuracy and DP by simply modifying predictions of a single base model according to different thresholds of the bias score. For Adult Census, the $\kappa(X)$ estimated over validation set is robust when evaluated over test set since the DPs for test set are either below the desired $\delta$'s or close to it. For COMPAS, the performance seems to be relatively low as one can see a relatively large drop in accuracy when DP is reduced. Although MBS still outperforms
\citetalias{zafar2017fairness}
in this case, it achieves worse performance than
\citetalias{jiang2019wasserstein}
and since the validation set is small (1056 instances), the $\kappa(X)$ estimated over it is not robust as there is a relatively big gap between DPs on test set and the specified $\delta$'s. The decline of performance on COMPAS is not surprising since COMPAS is a small dataset (with only 5278 instances), and the number of training examples is insufficient for reliable estimation of both $p(Y|X)$ and $p(A|X)$. 

\subsection{Experiments with EO as fairness criterion}
To evaluate the performance of MBS when the fairness criterion is EO, we again consider Adult Census and COMPAS datasets and the results are reported in Figures~\ref{fig_eo_adult}-\ref{fig_eo_compas}. The observation is similar to that in experiments with DP constraints. We again achieve better trade-off between accuracy and EO than Z17 
for both datasets. Although \citetalias{hardt2016equality} can also significantly reduce EO, similar to 
J19 
in the DP-based experiments, it is not able to adjust the balance between EO and accuracy and thus is less flexible than MBS. Both MBS and
\citetalias{zafar2017fairness}
achieve relatively low performance for small dataset COMPAS, which we again believe it is due to unreliable estimation of $p(Y,A|X)$ with small dataset. Similar to
J19
, H16
 assumes access to the sensitive in test, and thus will not be affected by unreliable inference of the test sensitive attribute.

\begin{table}[t]
\centering
\caption{Test accuracy (ACC) (\%) and EO (\%) on CelebA for all six combinations of target and sensitive attributes: (a, m), (a, y), (b, m), (b, y), (e, m), and (e, y) under different desired levels ($\delta$'s) of EO (\%) constraints. \textbf{Boldface} is used when MBS is better than \citetalias{park202fair} both in terms of accuracy and EO.}

\begin{adjustbox}{width=\textwidth}
\begin{tabular}{c  c c  c c c c c c c c c c } 
\hline
 &\multicolumn{2}{c}{(a, m)} & \multicolumn{ 2}{c}{(a, y)}  & \multicolumn{ 2}{c}{(b, m)} &\multicolumn{2}{c}{(b, y)} & \multicolumn{ 2}{c}{(e, m)}  & \multicolumn{ 2}{c}{(e, y)}  \\ 
 $\delta$ & ACC  & EO  & ACC  & EO  & ACC  & EO & ACC  & EO  & ACC  & EO  & ACC  & EO \\ 
 \hline
  $\infty$ & 81.9 & 23.8 & 82.0 & 24.7 & 84.1& 44.9 & 84.5& 21.0 & 85.2 & 19.0 & 85.6 & 13.7 \\
  10 & 80.9 & 6.7 & \textbf{79.7} & \textbf{10.9} & 83.5& 8.4 & 84.3 & 10.6 & 85.0& 8.2 & 85.1 & 9.5\\ 
  5 & \textbf{80.1} & \textbf{2.2} & 78.2 & 6.6 & \textbf{83.3} & \textbf{4.3} & \textbf{84.2} & \textbf{4.1}  & 84.7 & 5.6 & 85.1 & 5.4 \\ 
  1 & 79.3 & 4.0 & 76.9& 2.0 & 83.2 & 1.7 & 83.6 & 2.4 & 83.1& 1.7 &83.6 & 1.8 \\
   \citetalias{hardt2016equality} &  71.7& 3.8& 71.4 & 3.4 &  78.2 &  0.5 &   80.1 & 1.3 &  80.2 & 1.7 &  81.7 & 1.6 \\
  \citetalias{park202fair} &  79.1 & 6.5 & 79.1 & 12.4 &  82.9 & 5.0  &  84.1 & 4.8 &  83.4 & 3.0 &  83.5 & 1.6 \\
\hline
\label{table: eo_celeba1}
\end{tabular}
\end{adjustbox}
\end{table}

In addition, we evaluate MBS on a more challenging dataset, CelebA, following the same set-up as in \citetalias{park202fair}. Here we denote the target attributes ``Attractive", ``Big\_Nose", ``Bag\_Under\_Eyes" as ``a'', ``b'' and ``e'' respectively, and the sensitive attributes ``Male" and ``Young" as ``m'' and ``y'' respectively. The results are reported in Table~\ref{table: eo_celeba1}. MBS tends to achieve better trade-off than \citetalias{hardt2016equality}, 
whose accuracy is severely hurt across all 6 tasks. MBS is able to maintain high accuracy and meanwhile achieve competitive or even smaller EO. Furthermore, MBS consistently achieves better or competitive performance when compared with \citetalias{park202fair}. 
 To our knowledge, their method is one of the state-of-the-art methods for fair learning on CelebA. 
We additionally report validation metrics and standard error across 3 independent runs in Appendix \ref{appendix: tables}.

%% file: sections/conclusion.tex
To the best of our knowledge, we have for the first time characterized the Bayes optimal binary classifier under composite group fairness constraints, with a post-hoc modification procedure applied to an unconstrained Bayes optimal classifier. Our result applies to popular fairness metrics, such as DP, EOp, and EO as special cases. Based on this characterization, we propose a simple and effective post-processing method, MBS, which allows us to freely adjust the trade-off between accuracy and fairness. Moreover, MBS does not require test sensitive attribute, which significantly broadens its application in real-world problems where sensitive attribute is not provided during inference.

%% file: appendix/proof.tex
\subsection{Simple proof for the case of Demographic Parity}\label{simple_proof_for_dp_case}

For the sake of exposition, we first show a simple proof for the case of demographic parity. Observe that,
\begin{align*}
    Acc(\check{Y}) = & \textcolor{gray}{\text{(not  modifying term)}} \int (1 - \kappa(X)) p(Y = \hat{Y}| X) d\Pr(X) \\
    & \, + \textcolor{gray}{\text{(modifying term)}} \int \kappa(X) (1 -  p(Y = \hat{Y}(X)| X)) d\Pr(X) \\
    = & \, Acc(\hat{Y}) + \int \kappa(X) [ 1 - 2 p(Y = \hat{Y}(X) | X) ] d\Pr(X) \\
    = & \, Acc(\hat{Y}) - \int \kappa(X) \eta(X) d \Pr(X),
\end{align*}

Let us now write how the demographic parity changes after we { modify} the answers. We use notation $DP_{\pm}$ to denote a signed demographic parity that measures difference between positive output for groups $A = 0$ and $A = 1$, so that the actual $DP$ is the absolute value of $DP_{\pm}$. We have,
\begin{align*}
    DP_{\pm}(\check{Y}) &= \int \Pr(\check{Y} = 1 | X) \left[ \frac{p(A = 0 | X)}{\Pr(A = 0)} -\frac{p(A = 1 | X)}{\Pr(A = 1)} \right] d\Pr(X)  \\
    &= \int \left(\hat{Y} + \kappa(X) \{1 - 2 \hat{Y}\} \right) \left[ \frac{p(A = 0 | X)}{\Pr(A = 0)} -\frac{p(A = 1 | X)}{\Pr(A = 1)} \right] d\Pr(X) \\
    & = DP_{\pm}(\hat{Y}) - \int \kappa(X) \{2 \hat{Y} - 1\} \left[ \frac{p(A = 0 | X)}{\Pr(A = 0)} -\frac{p(A = 1 | X)}{\Pr(A = 1)} \right] d\Pr(X),
\end{align*}
here $DP_{\pm}(\hat{Y})$ corresponds to the parity of the original classifier $\hat{Y}$. W.l.o.g., assuming that $DP_{\pm}(\hat{Y})$ is positive (i.e., $A= 0$ is the advantaged group based on the unfair $\hat{Y}$), to decrease the DP maximally, we should sooner reject points with higher score
\[
    f(X) = \{2 \hat{Y} - 1\} \left[ \frac{p(A = 0 | X)}{\Pr(A = 0)} -\frac{p(A = 1 | X)}{\Pr(A = 1)} \right],
\]
It suggests when $\hat{Y}(X) = 1$, we should sooner {modify} the prediction (from 1 to 0) for $X$ that is more likely to come from the advantaged group favoured by the original unfair classifier, and when $\hat{Y}(X) = 1$, we should sooner {modify} the prediction (from 0 to 1) for $X$ that is more likely to come from the disadvantaged group based on the original unfair classifier. Similarly, if $DP_{\pm}^*$ is negative, we should {modify} those points with lowest score. Notice that we always should {modify} points with sign equal to that of $DP_{\pm}^*$, unless it is not possible, then we have reached ideal parity $DP_{\pm}(\kappa) = 0$. Recall that in order to reduce the accuracy the least, we should { modify} on the points with lowest confidence $\eta(X)$. The relative change then is quantified by the ratio $ s(X) = f(X) / \eta(X)$. Turns out, we can prove that the optimal rule correponds to $\kappa(X) = \Ind\{ s(X) > t \}$ (when $DP_{\pm}(\hat{Y}) >0$, otherwise we should look for $\kappa(X) = \Ind\{ s(X) < t \}$) by repeating the steps of famous Neyman-Pearson lemma, where we replace the probability of null and alternative hypothesis with $\eta(X)$, $f(X)$, respectively.

Indeed, by contradiction, assume there is a separate rule $\kappa'$ that achieves the same accuracy but smaller $DP_{\pm}$, so we have that
\[
    \int (\kappa'(X) - \kappa(X)) \eta(X) d\Pr(X) = 0, \;\;
    \text{and}
    \;\;
    \int (\kappa'(X) - \kappa(X)) f(X) d\Pr(X) < 0
\]
We assume that $f(X)$ is non-negative everywhere where both $\kappa(X)$ and $\kappa'(X)$ are non-negative, otherwise it corresponds to modifying a good answer on points from disadvantaged group. Let us denote $m(X) = (\kappa'(X) - \kappa(X)) \eta(X) $ and $r(X) = f(X)/ \eta(X)$. Then, we have that where $m(X) < 0$ we have $r(X) \geq s$ and where $m(X) > 0$ we have $r(X) < s$. So, we weight negative points with a smaller and positive score, hence contradiction. More, rigorously,
\begin{align*}
    \int m(X) r(X) d\Pr(X) & = \int m(X) r(X) \mathbbm{1}(m(X) > 0) d\Pr(X) \\
    & \phantom{=} + \int m(X) r(X) \mathbbm{1}(m(X) < 0) d\Pr(X) \\
    & \geq \int m(X) s \mathbbm{1}(m(X) > 0) d\Pr(X) + \int m(X) s \mathbbm{1}(m(X) < 0) d\Pr(X) \\ &= 0 \, .
\end{align*}

\subsection{Complete proof of Theorem~\ref{main_thm}}

First, let us expand the derivation in \EqRef{c_j_integral}. We have,
\begin{align*}
    C_{k}(\check{Y}) &= \int \Pr(\check{Y} = 1 | X) \left[ \frac{p(A_k = a_k | X)}{\Pr(A_k = a_k)} -\frac{p(A_k = b_k | X)}{\Pr(A_k = b_k)} \right] d\Pr(X) \\
    &= C_{k}^{*} - \int \kappa(X) \{2 \mathbbm{1}[\hat{Y} = 1] - 1\} \left[ \frac{p(A_k = a_k | X)}{\Pr(A_k = a_k)} -\frac{p(A_k = b_k | X)}{\Pr(A_k = b_k)} \right] d\Pr(X) \\
    &= C_{k}^{*} - \int \kappa(X) f_k(X) d\Pr(X),
\end{align*}
where we set $ C_{k}^* = C_k(\hat{Y}) $. Therefore, our optimization problem can be formulated as a linear problem on functions $\kappa(X)$
\begin{align*}
    Acc(\hat{Y}) - Acc(\check{Y}) = &\int \eta(X) \kappa(X) d\Pr(X) \rightarrow \min_{\kappa(X) \in [0, 1]},\\
    & \text{s. t.} \\
    &%
    \begin{aligned}
        \int f_k(X) \kappa(X) d\Pr(X) &\leq C_k^{*} + \delta \\
        -\int f_k(X) \kappa(X) d\Pr(X) &\leq - C_k^{*} + \delta
    \end{aligned} 
\end{align*}
To prove Theorem~\ref{main_thm}, it is left to apply the following technical lemma.

\begin{lemma}\label{lemma_linear_flipping_rule}
Suppose, we have functions $f_1(X), \dots, f_K(X), \eta(X)$, and a probability measure $P$ over $X \in \mathcal{X}$. Consider the optimization problem over measurable functions $\kappa(X) $ taking values in $[0, 1]$,
\[
    \langle \kappa, \eta \rangle \rightarrow \min, \qquad \text{s.t.} \qquad \langle \kappa, f_k \rangle \leq b_k, \;\; k = 1, \dots, K ,
\]
where $\langle f, g \rangle = \int f(X)g(X) dP(X)$. Suppose, the following conditions hold:
\begin{enumerate}[A.]
    \item all functions $f_{k}, \eta $ are from $L_{2}(P)$;
    \item there is a strictly feasible solution, in the sense that there is a function $\kappa'(X) \in [0, 1]$ such that $ \langle \kappa', f_{k} \rangle < b_{k} $;
    \item for any $ z \in \R^{k}_{+} $, we have that $P(\sum_{k} z_k f_k(X) + \eta(X) = 0) = 0$. Notice that this is a weaker version of the condition that  $ (f_1 / \eta, \dots, f_{K} / \eta) $ has continuous distribution.
\end{enumerate} Then, there is $z \in \R^{K}_{+}$ and an optimal solution of the
\[
    \kappa(X) = \mathbf{1}\left[ \sum_{k} z_{k} f_{k}(X) > \eta(X) \right ] \, .
\]
Furthermore, $z \in \R^{k}_{+}$ can be any solution to the problem $ \min_{z \geq 0} b^{\T} z + \int (z^{\T} F(x) + \eta(x))_{+} dP(x) $.
\end{lemma}

Notice, that thanks to $\delta > 0$, the restrictions can be met with strict inequalities by the {modification} rule $ \kappa(X) = \mathbf{1}[p(Y = 1 | X) < 0.5]$ which corresponds to the constant classifier $ \check{Y} = 1 $. Furthermore, we note that we have in total $2K$ restrictions, one for each of functions $f_k$, $-f_k$), and the resulting vector $z$ in \EqRef{optimal_kappa_from_theorem} therefore can have negative values, unlike the vector $z$ in the above lemma.

\begin{remark}
\cite{zeng2022bayes} refer to the \emph{generalized Neyman-Pearson lemma}, which looks very similar to the lemma above. In their formulation however, it is required that such $z_1, \dots, z_k$ exist that turn all inequality constraints into equalities. We avoid this requirement by using linear programming duality. In general, the optimal $z_1, \dots, z_k$ do not have to satisfy all constraints as equality, which extends the range of cases where our lemma can be applied.
\end{remark}

\begin{proof}[Proof of Lemma~\ref{lemma_linear_flipping_rule}]
Consider an operator $F: L_2(P) \mapsto \R^{K}$ such that $ F \kappa = (\langle f_1, \kappa \rangle, \dots, \langle f_K, \kappa \rangle)^{\T}$. Below we write that a vector $x$ satisfies $x \geq 0$ if all its coordinates are non-negative, i.e. $x \in \R_{+}^{d}$ for some $d$. With some abuse of notation we will also say that $ z^{\T} F = \sum_{k} z_k f_k $. Denote $S_{+}$ the set of $L_2$ functions with pointwise non-negative values. Our problem reads as follows,
\[
    \max_{\kappa} - \langle \eta, \kappa \rangle \qquad \text{s.t.}\qquad \kappa \in S_{+}, \;\; 1 - \kappa \in S_{+}, \;\; F\kappa  \leq b \,.
\]
Consider the dual problem,
\begin{equation}\label{dual_problem}
    \min_{z, \lambda}  z^{\T} b + \langle \lambda, 1 \rangle,
    \qquad
    \text{s.t.}
    \qquad
    z \in \R^{K}_{+}, \lambda \in S_{+},
    z^{\T} F + \lambda + \eta \in S_{+} \,.
\end{equation}
We have the duality property, for any feasible $\kappa $ to primal and $z, \lambda$ to dual, we have
\begin{align*}
    - \langle \eta, \kappa  \rangle \leq z^{\T} F \kappa  + \langle \lambda, \kappa  \rangle \leq z^{\T} b + \langle \lambda, 1 \rangle .
\end{align*}
In the linear programming these inequalities are called \emph{slackness condition}.
Assume for a moment, that strong duality holds, i.e. there is $\kappa $ and $z, \lambda$ such that all inequalities turn into equalities. Then, both pairs are $\kappa$ and $z, \lambda$ are optimal for their linear programs. Furthermore,
\[
    \langle \lambda, 1 - \kappa \rangle = 0,
    \qquad 
    \langle \kappa, z^{\T} F + \eta + \lambda \rangle = 0
\]
Furthermore, it is straightforward to see that the optimal solution should satisfy almost everywhere $\lambda = (F^{\T}z +  \eta)_{-} $, where for a function $g(X)$ we denote its negative part $g_{-}(X) = \max(0, -g(X))$. Simply observe that we must minimize each $\lambda(X)$ independently, while satisfying the condition $ z^{\T} F(X) + \eta(X) + \lambda(X) \geq 0 $.
Therefore, we have that the set $\{ X: z^{\T} F + \eta + \lambda > 0\} $ is the same as $\{ X: z^{\T} F + \eta > 0\}$, up to a difference of probability $0$. So, we can say that $ z^{\T} F + \eta > 0 $ yields $ \kappa = 1$ $P$-a.s. and $ z^{\T} F + \eta < 0 $ yields $ \kappa = 0$ $P$-a.s. Assuming $ P(z^{\T} F + \eta = 0) = 0$, we have that there is an optimal solution that has a form $ \kappa = \Ind\{ z^{\T} F + \eta > 0\} $. 

Now let us show the strong duality property. Consider the value
\(
    v^* = \min_{z \geq 0} z^{\T} b + \langle (z^{\T}F + \eta)_{-}, 1 \rangle,
\)
which is the optimal value in the dual problem, thanks to the identity for the optimal $\lambda$ given $z$. Thanks to its closed form, we show that it is stable w.r.t. the perturbations in the inputs $F, \eta$, so that we can reduce our problem to finite dimensional linear programming (LP), where strong duality is known to hold.

\textbf{Step 1 (bounded $z$).} First let us show that any optimal $z$ is bounded. Since for the primal problem  the feasible set has non-empty interior, we have that for every $a \in \R_{+}^{K}$
\[
    a^{\T} b + \langle (a^{\T} F)_{-}, 1 \rangle > 0,
\]
and notice that this is a continuous function of $a \in \R_{+}^{K}$ (follows from $f_k \in L_1$). Therefore, the following number is strictly positive
\begin{equation}\label{z_f_equation_}
    \mathcal{Z}_{F} := \inf_{\| a \| = 1, a \geq 0}  a^{\T} b + \langle (a^{\T} F)_{-}, 1 \rangle > 0 \, .
\end{equation}
Let us show by contradiction that any optimal solution $ z \geq 0 $ to dual problem satisfies $ \| z \| \leq \langle |\eta|, 1 \rangle / \mathcal{Z}_{F} $. Assume $ \| z \| > \langle |\eta|, 1 \rangle / \mathcal{Z}_{F} $. We have that for $ z = L a$ with $\| a \| = 1$, 
\[
    L a^{\T} b + \langle (L a^{\T} F + \eta )_{-}, 1 \rangle \geq L (a^{\T} b + \langle (a^{\T} F) , 1 \rangle) - \langle \eta_{+} , 1 \rangle > \langle \eta_{-} , 1 \rangle
\]
as long as  $L > \langle |\eta|, 1 \rangle / \mathcal{Z}_{F} $. {Since} the  objective value $\langle \eta_{-} , 1 \rangle$ can be achieved with $ z = 0 $ {in the dual problem}, so an optimal $z$ must have a norm smaller or equal to $ \langle |\eta|, 1 \rangle / \mathcal{Z}_{F}  $.

\textbf{Step 2 (discrete approximation).}
Next, consider step approximations of $F$ and $\eta$ in the following form. Select some partition $ I_1 \cup \dots \cup I_{M} = \mathcal{X} $, and set $ \tilde{f}_{k}(X) = \sum_{m} \tilde{f}_{km} \Ind\{X \in I_{m}\} $ , $ \tilde{\eta}(X) = \sum_{m} \tilde{\eta}_{m} \Ind\{X \in I_{m}\} $, in a way that $ \| \eta - \tilde{\eta}\|_{L_2} \leq \eps $ and $ \| f_k - \tilde{f}_k\|_{L_2} \leq \eps $. The fact that $f_k, \eta \in L_1$ is enough to find such approximation, however, since we require $ f_{k}, \eta  \in L_{2}$ we can also assume that $ \tilde{\eta}_{k} = \E[\eta(X) | X \in I_{m}] $, $ \tilde{f}_{km} = \E[f_{k}(X) | X \in I_{m}] $, where each $ I_{m} $ is assumed to have non-zero probability. Let us consider,
\[
    \tilde{v} = \min_{z \geq 0} z^{\T} b + \langle (z^{\T}\tilde{F} + \tilde{\eta})_{-}, 1 \rangle
\]
An optimal $ \tilde{z} $ to this problem is bounded by $ (\langle |\eta|, 1 \rangle + \eps) / (\mathcal{Z}_{F} - k \eps) \leq  2 \langle |\eta|, 1 \rangle / \mathcal{Z}_{F} $ for sufficiently small $ \eps $. We therefore can assume that both problems are minimized in $ [0, D]^{K}$ instead of $ \R^{K}_{+} $ for some $D$ that does not depend on (small enough) $\eps$.

We have then (denoting by $\tilde{y}$ an optimal solution for $\tilde{v}$) that
\begin{align*}
    \tilde{v} = \tilde{z}^{\T} b + \langle (\tilde{z}^{\T} \tilde{F} + \tilde{\eta})_{-}, 1 \rangle \geq \tilde{z}^{\T} b + \langle (\tilde{z}^{\T} F + \tilde{\eta})_{-}, 1 \rangle - \langle |\tilde{z}^{\T} (F - \tilde{F})|, 1\rangle  \geq v^{*} - DK\eps \, .
\end{align*}
Similarly, $ v^{*} \geq \tilde{v} - DK\eps $. Let us look at the approximation of the primal problem. Let $ u^{*} $ and $\tilde{u}$ correspondingly be the optimal values of the objective in the original and discrete problem. Let us take the discrete function $ \tilde{\kappa}(X) = \sum_{m} \tilde{\kappa}_{m} \Ind\{ X \in I_m\} $, where $\tilde{\kappa}_{m}$ is the elements of the vector that solves the discrete problem. Then,
\(
    u^{*} \geq - \langle \tilde{\kappa}, \eta \rangle = - \langle \tilde{\kappa}, \eta \rangle = \tilde{u},
\)
where we used the fact that $ \tilde{\eta}$ averages $ \eta $ over each interval $I_{m}$. Thanks to the strong duality of LP, for the discrete problem we have that $ \tilde{u} = \tilde{v}$. Thus, we have shown that $ u^{*} \geq \tilde{u} = \tilde{v} \geq v^{*} - DK \eps  $, while $u^{*} \leq v^{*}$ is already guaranteed by weak duality. Sending $\eps$ to zero proves the required strong duality property. 
\end{proof}

%% file: appendix/algorithm.tex
\subsection{Algorithms}\label{appendix: alg}

Here we present the detailed algorithms we used in our experiments for DP and EO, particularly, for the choice of linear rule. In each case, we assume that a validation set $ \{(X_i, Y_i, A_i)\}_{i = 1}^{N_{val}}$ is given, based on which we choose the rule satisfying empirical fairness restrictions, whilst maximizing the empirical accuracy. 

For the case of DP restrictions, we assume that we are given models $ \hat{p}(Y|X), \hat{p}(A|X) $, as well as evaluated probabilities $ \widehat{\Pr}(A=0), \widehat{\Pr}(A = 1) $, which we can either evaluate on a bigger training dataset (in case we have such access), or on the validation dataset. After we calculate the corresponding scores $\hat{s}(X_i)$, we consider rearrangement $ \hat{s}(X_{i_1}) \geq \dots \geq \hat{s}(X_{i_{N_{val}}}) $ and test the thresholds $ t = \hat{s}(X_{i_j}) $ one after the other. When we move to the next candidate $ t = \hat{s}(X_{i_{j + 1}}) $ we only need to spend $O(1)$ time updating the current empirical DP and accuracy. See detailed procedure in Algorithm~\ref{alg: dp}.

For the case of EO restrictions, we assume that we are given models $ \hat{p}(Y|X), \hat{p}(Y, A|X) $, which are two functions that do not necessarily agree as probability distributions. We also assume that we are given evaluated probabilities $ \widehat{\Pr}(Y = y, A=a), a, y \in \{0, 1\}$. Then, we calculate the corresponding scores $\hat{s}_{0, 1}(X_i)$. In order to choose the linear rule of the form $ \kappa(X) = \Ind\{ w_0 \hat{s}_0(X) + w_1 \hat{s}_1(X) > t\} $ we consider two strategies. In the first one, we observe that we can restrict the search only to lines that pass through two validation points, which exhaust the search of all possible pairs of accuracy and EO on validation. This, however, requires $O(N^3)$ to evalueate, so we consider an option to $M$ sub-samples and only consider the corresponding $M(M-1)/2$ lines passing through every two of them. See pseudo-code in Algorithm~\ref{alg: eo}.

For another option, we fix $K$ possible directions $ (w_0, w_1) \in \R^{2}$, such that $w_0^2 + w_1^2 = 1$, and then for each of them consider the score $ \hat{s}(X) = w_0 \hat{s}_0(X) + w_1 \hat{s}_1(X) $. This way we only need to choose the optimal thresholds $c$, similar to the case of DP. By doing so for each of the $K$ directions, we can approximate the best line in $O(KN_{val}\log N_{val})$ time (which includes sorting the score values). See detailed procedure in Algorithm~\ref{alg: eo2}.


\begin{algorithm}
    \caption{Algorithm for DP}\label{alg: dp}
    \begin{algorithmic}[1]
    \Require 
    $\delta$: desired constraint for DP, 
    
    $\{(X_i, Y_i, A_i)\}_{i=1}^{N_{val}}$: validation set, 
    
    $\hat{p}(Y|X)$, $\hat{p}(A|X)$,

    evaluated $\widehat{\Pr}(A= 0)$, $\widehat{\Pr}(A = 1)$
    
    \Ensure Threshold $t^\ast$ for modification rule $\kappa(X) = \Ind\{ \hat{s}(X) > t^\ast \}$:

     \State Compute $\{\hat{Y}_i=\Ind\{ \hat{p}(Y=1 |X_i) > 0.5\}\}_{i=1}^{N_{val}}$

    \State $\mathtt{acc} = \widehat{Acc}(\hat{Y})$
    \State $ \mathtt{counts}[a] = $ calculate frequency of $\hat{Y} = 1$ in two groups $A = 0, 1 $
    
    \State Compute $\{\hat{s}(X_i)\}_{i=1}^{N_{val}}$ based on $\hat{p}(Y|X)$, $\hat{p}(A|X)$, and $\widehat{Pr}(A)$

    \State Compute argsort $ IDX = [i_1, \dots, i_{N_{val}} ]$ in ascending order, such that $ \hat{s}(X_{i_j}) \geq \hat{s}(X_{i_{j + 1}}) $
    \State Best accuracy $ACC_{\max} = 0$
    
    \For{$j$ in  $IDX$}
    \State set candidate $t = \hat{s}(X_{j})$
    \State $\mathtt{acc} = \mathtt{acc} + N_{val}^{-1}( 1 - 2 \Ind\{ \hat{Y}_j = Y_j\})$ \Comment{Update accuracy}
    \State $\mathtt{counts}[A_{j}] = \mathtt{counts}[A_{j}] + N_{val}^{-1}(1 - 2 \hat{Y}_j)$ \Comment{Update counts for group $A = A_j$} 
    \State $\mathtt{dp} = \left|\mathtt{counts}[0] / \widehat{\Pr}(A = 0) - \mathtt{counts}[1] / \widehat{\Pr}(A = 1)\right|$
    \If{$\mathtt{dp} \leq \delta $ and $\mathtt{acc} > ACC_{\max}$}
        \State $t^{*} = t$ \Comment{Update best candidate}
        \State $ ACC_{\max} = \mathtt{acc}$
    \EndIf
    \EndFor
    \State \textbf{return} $t^*$
    \end{algorithmic}
\end{algorithm}

\begin{algorithm}
    \caption{Algorithm for EO}\label{alg: eo}
    \begin{algorithmic}[1]
    \Require 
    $\delta$: desired constraint for EO, 

    $M$: the number of random samples from the validation set
    
    $\{(X_i, Y_i, A_i)\}_{i=1}^{N_{val}}$: validation set,
    
    $\hat{p}(Y|X)$, $\hat{p}(Y,A|X)$,
    
    evaluated $ \widehat{\Pr}(Y = y, A=a), a, y \in \{0, 1\}$
    
    \Ensure Linear modification rule $\kappa(X)$

    \State Compute $\{\hat{Y}_i=\Ind\{ \hat{p}(Y=1 |X_i) > 0.5\}\}_{i=1}^{N_{val}}$
    
    \State Compute $\{[\hat{s}_0(X_i), \hat{s}_1(X_i)]^\top\}_{i=1}^{N_{val}}$ based on $\hat{p}(Y|X)$, $\hat{p}(Y,A|X)$ and $\widehat{\Pr}(Y, A)$

    \State Randomly sample $M$ instances from the validation set: $\{(X_m', Y_m')\}_{m=1}^M$

    \State Construct the set of linear rules to search with index: \Comment{size: $\frac{M(M-1)}{2}$} \\
    \[
        \mathcal{S}_{M} = \left\{ (k, l) | k < l, \;\; k, l = 1, \dots, M \right\}
    \]
    
    \State Best accuracy $ACC_{\max} = 0$
    \For{$(k,l)$ in $\mathcal{S}_{M}$}
        \State \(\kappa^{(k,l)}(X)=\mathbf{1}\left\{ \frac{\hat{s}_1(X)-\hat{s}_1(X_k')}{\hat{s}_1(X_k')-\hat{s}_1(X_l')}>\frac{\hat{s}_0(X)-\hat{s}_0(X_k')}{\hat{s}_0(X_k')-\hat{s}_0(X_l')} \right\}\)
        \For{$i=1, \cdot\cdot\cdot, N_{val}$}  
            \If{$\kappa^{(k,l)}(X_i)=1$}
                \State $\check{Y}_i = 1- \hat{Y}_i$ \Comment{Modify when $\kappa^{(k,l)}(X_i)=1$}
            \Else{}
                \State $\check{Y}_i = \hat{Y}_i$ \Comment{Do not modify when $\kappa^{(k,l)}(X_i)=0$}
            \EndIf
        \EndFor
        \State Compute validation $ACC(\kappa^{(k,l)})$ and $EO(\kappa^{(k,l)})$ based on $\check{Y}$
        \If{$EO(\kappa^{(k,l)}) \leq \delta$ and $ACC(\kappa^{(k,l)})>ACC_{max}$}
            \State $\kappa(X)=\kappa^{(k,l)}(X)$,
            $ACC_{max}=ACC(\kappa^{(k,l)})$
        \EndIf
    \EndFor
    \State \textbf{return} $\kappa(X)$
    \end{algorithmic}
\end{algorithm}

\begin{algorithm}
    \caption{Alternative algorithm for EO}\label{alg: eo2}
    \begin{algorithmic}[1]
    \Require 
    $\delta$: desired constraint for EO, 
    
     $\{(X_i, Y_i, A_i)\}_{i=1}^{N_{val}}$: validation set,
    
    $\hat{p}(Y|X)$, $\hat{p}(Y,A|X)$,
    
    evaluated $ \widehat{\Pr}(Y = y, A=a), a, y \in \{0, 1\}$
    
    \Ensure Threshold $t^\ast$ and direction $(w_0^\ast, w_1^\ast)$ for modification rule $\kappa(X) =  \Ind\{ w_0^*\hat{s}_0(X) + w_1^* \hat{s}_1(X) > t^*\}$:

     \State Compute $\{\hat{Y}_i=\Ind\{ \hat{p}(Y=1 |X_i) > 0.5\}\}_{i=1}^{N_{val}}$

    \State $ \mathtt{counts}[y, a] = $ calculate frequency of $\hat{Y} = 1$ in all four groups $(Y, A) = (y, a) $
    
    \State Compute $\{(\hat{s}_0(X_i),\hat{s}_1(X_i)) \}_{i=1}^{N_{val}}$ based on $\hat{p}(Y|X)$, $\hat{p}(Y, A|X)$, and $\widehat{\Pr}(Y, A)$
    \State Fix set of directions $ \mathcal{W}_K =  \{ (w_0^{(k)}, w_1^{(k)}) \}_{k = 1}^{K} \subset \R^{2} $, $\| (w_0^{(k)}, w_1^{(k)})\|_2 = 1$
    \State Best accuracy $ACC_{\max} = 0$
    \For{$(w_0, w_1)$ in $\mathcal{W}_K$}
     \State $\mathtt{acc} = \widehat{Acc}(\hat{Y})$
    \State Compute scalar scores $ \hat{s}(X_i) = w_0 \hat{s}_0(X_i) + w_1 \hat{s}_1(X_i)$
    \State Compute argsort $ IDX = \{i_1, \dots, i_{N_{val}} \}$ in ascending order, such that $ \hat{s}(X_{i_j}) \geq \hat{s}(X_{i_{j + 1}}) $
    
    \For{$j$ in  $IDX$}
        \State set candidate $t = \hat{s}(X_{j})$
        \State $\mathtt{acc} = \mathtt{acc} + N_{val}^{-1}( 1 - 2 \Ind\{ \hat{Y}_j = Y_j\})$ \Comment{Update accuracy}
        \State $\mathtt{count}[Y_j, A_j] = \mathtt{count}[Y_j, A_j] + N_{val}^{-1}(1 - 2 \hat{Y}_j)$ \Comment{Update counts for group $(Y_j, A_j)$}
        \State $\mathtt{eo0} = \left|\mathtt{count}[0, 0] / \widehat{\Pr}(Y = 0, A = 0) - \mathtt{count}[0, 1] / \widehat{\Pr}(Y = 0, A = 1)\right|$
        \State $\mathtt{eo1} = \left|\mathtt{count}[1, 0] / \widehat{\Pr}(Y = 1, A = 0) - \mathtt{count}[1, 1] / \widehat{\Pr}(Y = 1, A = 1)\right|$
        \State $\mathtt{eo} = \max(\mathtt{eo0}, \mathtt{eo1})$
        \If{$\mathtt{eo} \leq \delta $ and $\mathtt{acc} > ACC_{\max}$}
            \State $ ACC_{\max} = \mathtt{acc}$ \Comment{update best candidate}
            \State $t^{*} = t$
            \State $(w_0^*, w_1^*) = (w_0, w_1)$
        \EndIf
        \EndFor
    \EndFor
    \State \textbf{return} Modification rule $\kappa(X) = \Ind\{ w_0^*\hat{s}_0(X) + w_1^* \hat{s}_1(X) > t^*\}$
    \end{algorithmic}
\end{algorithm}

\label{appendix: pick_line}

%% file: appendix/ablation.tex
\label{appendix: ablation}
Here we consider two ablation studies to examine the robustness of our method when the estimated conditional distributions are inaccurate. We demonstrate it with DP as fairness criterion, and we report the mean and standard error of the results from 3 independent runs.

\subsection{Corrupted $\hat{p}(Y|X)$ and $\hat{p}(A|X)$}
\begin{figure}[htbp]
    \centering
    \begin{subfigure}{0.49\textwidth}
        \includegraphics[width=\textwidth]{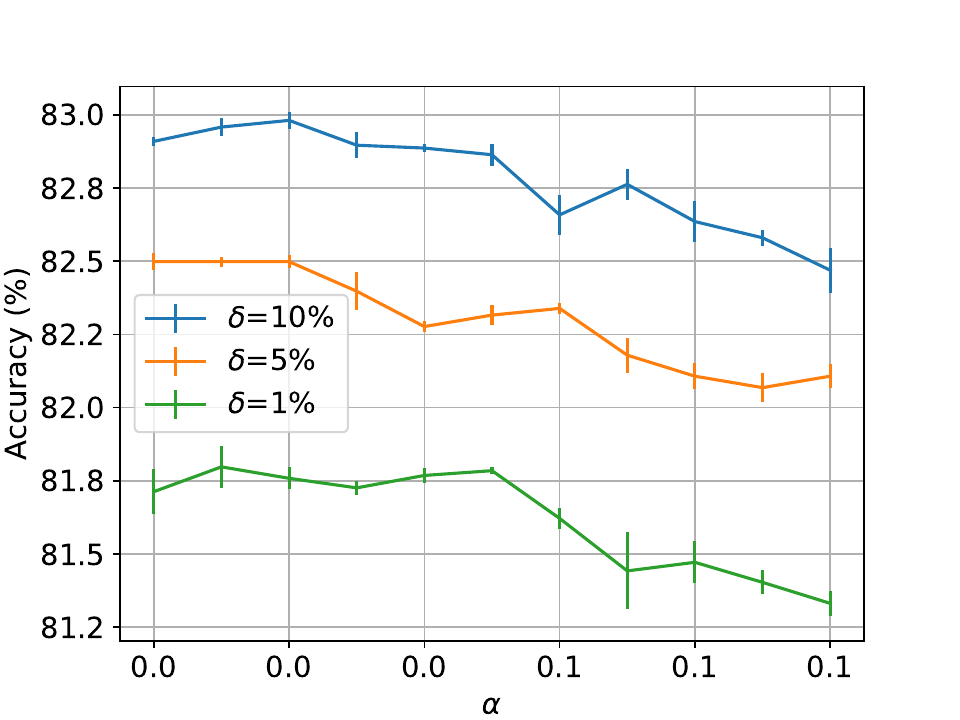}
        \caption{Accuracy ($\check{Y}$) (\%) vs $\alpha$}
    \end{subfigure}
    \begin{subfigure}{0.49\textwidth}
        \includegraphics[width=\textwidth]{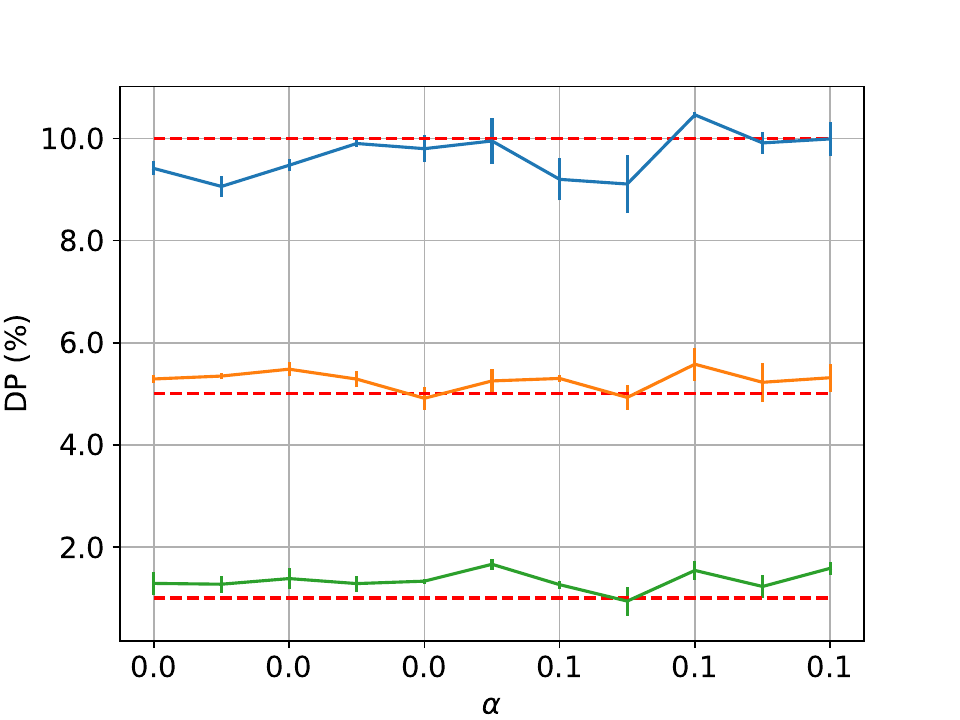}
        \caption{DP (\%) vs $\alpha$}
    \end{subfigure}
    \caption{Test set performance of our method with corrupted $\hat{p}(Y|X)$ and $\hat{p}(A|X)$ on Adult Census dataset.}
    \label{fig: ablation1_adult}
\end{figure}

\begin{figure}[htbp]
    \centering
    \begin{subfigure}{0.49\textwidth}
        \includegraphics[width=\textwidth]{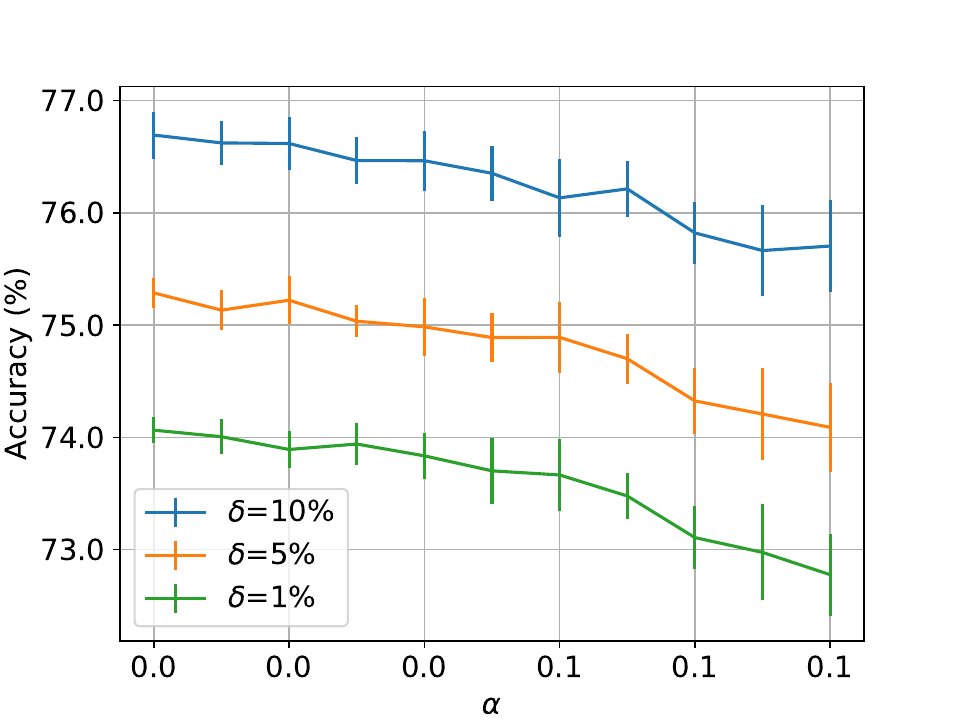}
        \caption{Accuracy ($\check{Y}$) (\%) vs $\alpha$}
    \end{subfigure}
    \begin{subfigure}{0.49\textwidth}
        \includegraphics[width=\textwidth]{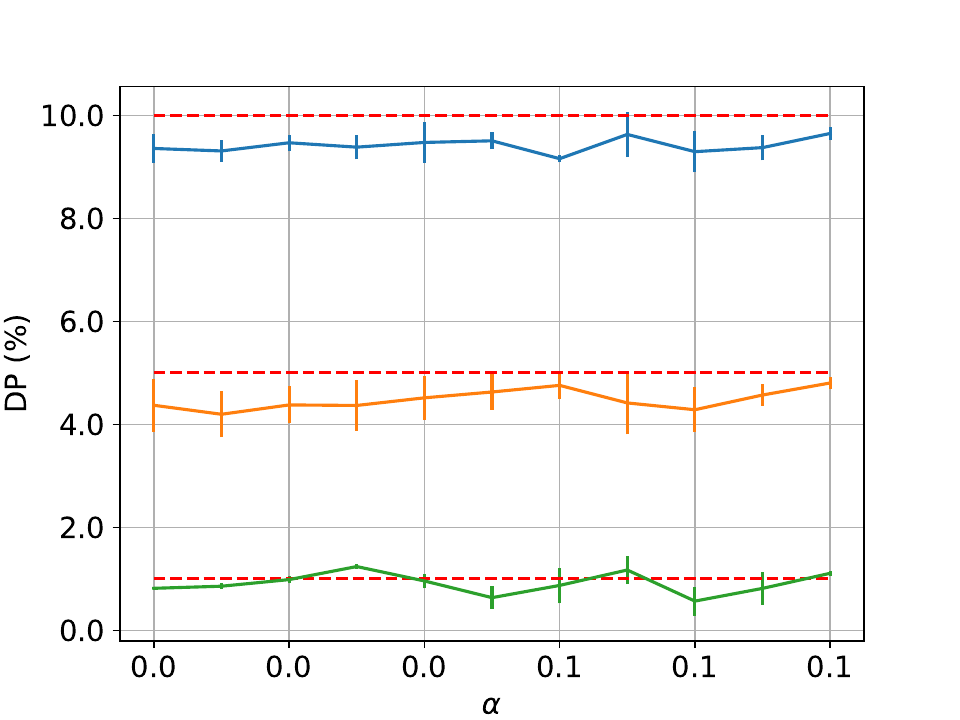}
        \caption{DP (\%) vs $\alpha$}
    \end{subfigure}
    \caption{Test set performance of our method with corrupted $\hat{p}(Y|X)$ and $\hat{p}(A|X)$ on CelebA dataset (target: ``Attractive'', sensitive attribute: ``Male'').}
    \label{fig: ablation1_celeba}
\end{figure}

In the first ablation study, we consider manually adding noise $\epsilon$ to $\hat{p}(Y|X)$ and $\hat{p}(A|X)$. For each $X$, we sample $\epsilon$ from a uniform distribution $Unif(-\alpha, 2\alpha)$, with $\alpha$ controlling the intensity of the corruption ($\alpha$ is chosen to be $0, 0.01, \dots, 0.1$). In Figures~\ref{fig: ablation1_adult} and~\ref{fig: ablation1_celeba}, we plot the 
post-processed test target accuracy ($\check{Y}$) and post-processed test DP after selecting the modification threshold on validation set based on corrupted $\hat{p}(Y|X)$ and $\hat{p}(A|X)$ for Adult Census and CelebA (with target ``Attractive'' and sensitive attribute ``Male'') respectively. Across all corruption intensity $\alpha$, test DP still tends to be below or fluctuate around its desired constraints $\delta$'s, while we only incur a minor drop in test target accuracy (within 1\% across most $\alpha$'s). This demonstrates the robustness of our post-processing algorithm even when the estimated conditional distributions do not match the ground-truth well.

\subsection{Strongly regularized $\hat{p}(A|X)$}
\begin{figure}[htbp]
    \centering
    \begin{subfigure}{0.32\textwidth}
        \includegraphics[width=\textwidth]{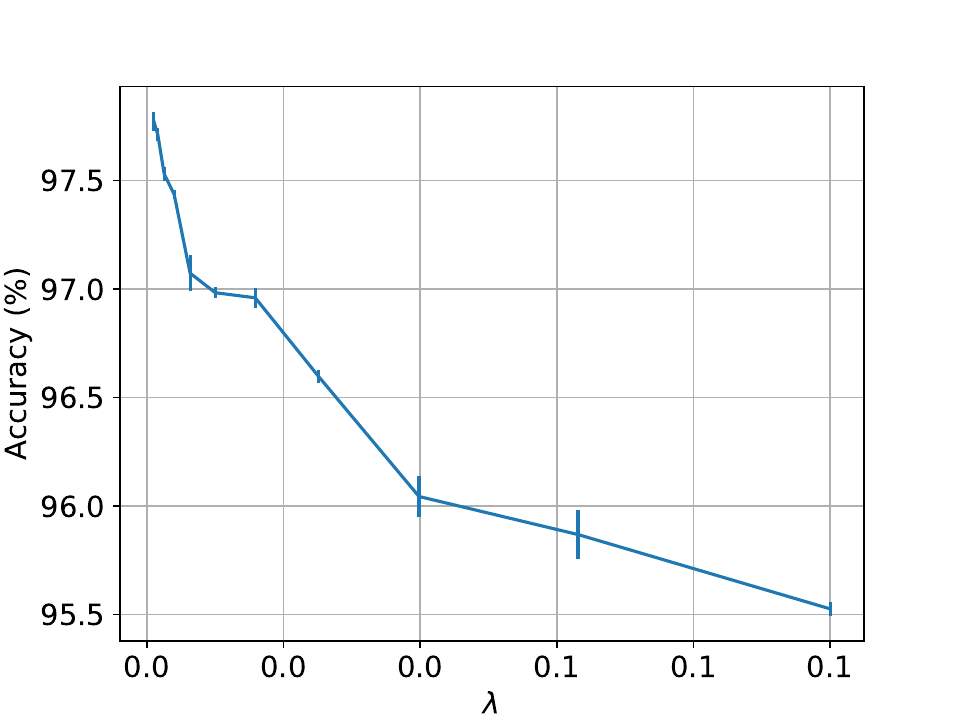}
        \caption{Accuracy ($\hat{A}$) (\%) vs $\lambda$}
    \end{subfigure}
    \begin{subfigure}{0.32\textwidth}
        \includegraphics[width=\textwidth]{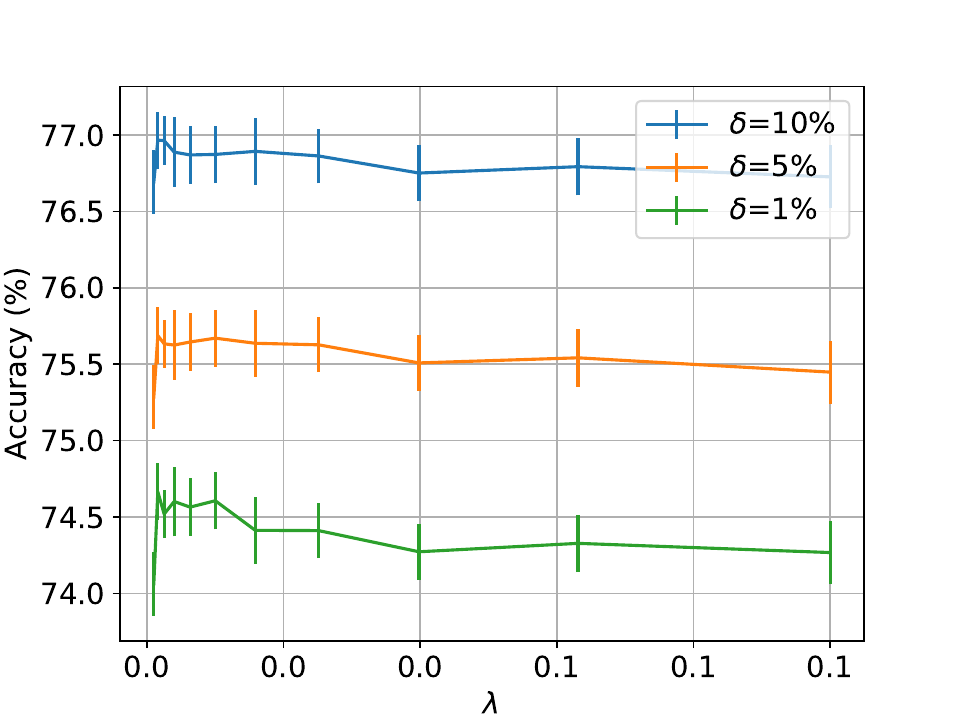}
        \caption{Accuracy ($\check{Y}$) (\%) vs $\lambda$}
    \end{subfigure}
    \begin{subfigure}{0.32\textwidth}
        \includegraphics[width=\textwidth]{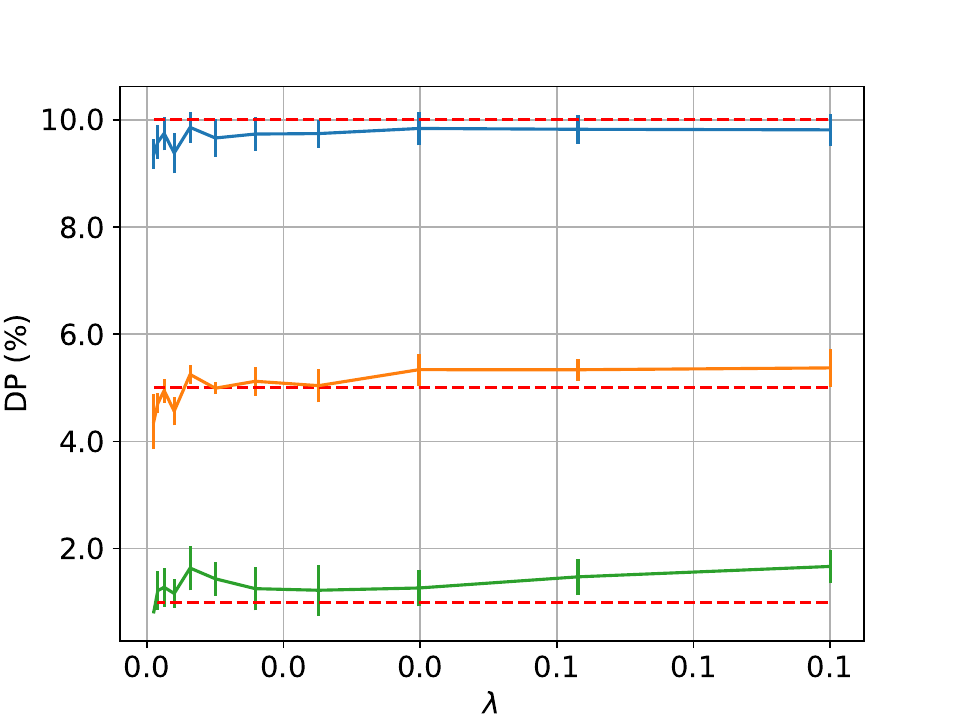}
        \caption{DP (\%) vs $\lambda$}
    \end{subfigure}
    \caption{Test set performance of our method based on $\hat{p}(A|X)$ trained with different weight decay $\lambda$ on CelebA dataset (target: ``Attractive'', sensitive attribute: ``Male'').}
    \label{fig: ablation2}
\end{figure}

In the second ablation study, we reduce the discrimination power of the sensitive classifier by training $\hat{p}(A|X)$ with stronger regularization. Specifically, for CelebA (target ``Attractive'' and sensitive attribute ``Male''), we train multiple sensitive classifiers with different weight decay $\lambda$ increasing from 0.001 to 0.1. Figure \ref{fig: ablation2} plots the test accuracy of sensitive model $\hat{p}(A|X)$, post-processed test target accuracy ($\check{Y}$), and post-processed DP after modification against $\lambda$ respectively. One can see that as $\lambda$ increases, the sensitive accuracy given by $\hat{p}(A|X)$ drops. However, with less accurate $\hat{p}(A|X)$, post-processed target accuracy ($\check{Y}$) and DP after modification does not seem to be noticeably hurt. Interestingly, when $\lambda$ increases moderately, although $\hat{p}(A|X)$ becomes less accurate, the post-processed target ($\check{Y}$) accuracy can even be improved. This further confirms the robustness of our post-processing modification algorithm.

\subsection{Zero-shot CLIP as auxiliary model}\label{sec:clip}
\begin{table}[htbp]
\centering
\caption{Experiment with CLIP: Accuracy (\%) and Equalized Odds (EO) (\%) on CelebA with target attribute ``Attractive'' and sensitive attributes ``Male''. Main unconstrained model $\hat{p}(Y|X)$ is trained on CelebA, while auxiliary model $\hat{p}(Y, A|X)$ is evaluated from CLIP's similarity to text prompts.}
 \begin{tabular}{c  c  c c   } 
 \hline
 & $\delta$ & ACC & EO \\  
\hline
  \multirow{ 5}{*}{Validation} 
  & $\infty$ & 81.1 & 27.8\\
   &  10 &  81.0 &  10.0 \\
  &  5 &  80.4 &  5.0 \\
  &  1 &  80.0 &  1.0 \\
  \hline
  \multirow{ 6}{*}{Test}
  & $\infty$ & 81.6 & 24.1\\
  &  10 &  80.5 & 6.2 \\
    &  5 &  79.9 & 1.0 \\
    &  1 &  79.1 & 5.4 \\
\hline
\label{table: eo_clip}
\end{tabular}
\end{table}
Here, we consider for CelebA the classification problem with target ``Attractive'' and sensitive attribute ``Male'', with EO constraints imposed. We train only the target model $\hat{p}(Y |X )$ on the original CelebA data, while the pair $\hat{p}(Y, A|X)$ is evaluated from CLIP. Namely, we consider the following prompts,
\begin{align*}
    \mathtt{prompt}\_{11} &= \text{``a photo of Attractive Male''}\\
    \mathtt{prompt}\_{10} &= \text{``a photo of Attractive Female''}\\
    \mathtt{prompt}\_{01} &= \text{``a photo of Unattractive Male''}\\
    \mathtt{prompt}\_{01} &= \text{``a photo of Unattractive Female''}
\end{align*}
For a given prompt and image $X$, let $sim(X, \mathtt{prompt})$ denotes the CLIP's similarity. We consider the following probabilities,
\[
    \hat{p}(Y = i, A = j|X ) = \frac{\exp(sim(X, \mathtt{prompt}\_{ij}))}{\sum_{i, j \in \{0, 1\}} \exp(sim(X, \mathtt{prompt}\_ij)}
\]
We then feed these two models into Algorithm~\ref{alg: eo2}. Results are reported in Table~\ref{table: eo_clip}.

%% file: appendix/tables.tex
We present numerical results (mean±standard error) for all experiments on CelebA in tables.


\label{appendix: tables}

\begin{table}[htbp]
\centering
\caption{Accuracy (\%) and Equalized Odds (EO) (\%) on CelebA with target attribute ``Attractive'' and sensitive attributes ``Male'' under different desired levels ($\delta$'s) of EO (\%) constraints. \textbf{Boldface} is used when MBS is better than \citet{park202fair} both in terms of accuracy and EO.}
 \begin{tabular}{c  c  c c   } 
\hline
 & $\delta$ & ACC & EO \\ 
 \hline
  \multirow{ 5}{*}{Validation} 
  & $\infty$ & 81.3$\pm$0.1 & 26.8$\pm$1.0\\
  & 10 & 80.8$\pm$0.1 & 9.9$\pm$0.0\\
  &5 & 80.2$\pm$0.2 & 5.0$\pm$0.0\\
  &1 & 79.6$\pm$0.2 & 0.7$\pm$0.2\\
  & \citet{hardt2016equality} &  71.3$\pm$0.1 & 0.5$\pm$0.1\\
  \hline
  \multirow{ 6}{*}{Test}
  & $\infty$ & 81.9$\pm$0.3 & 23.8$\pm$0.8\\
  & 10 & 80.9$\pm$0.2 & 6.7$\pm$0.6\\
  &5 & \textbf{80.1$\pm$0.3} & \textbf{2.2$\pm$0.5}\\
  &1 & 79.3$\pm$0.3 & 4.0$\pm$0.4 \\
  & \citet{hardt2016equality} & 71.7$\pm$0.3 & 3.8$\pm$0.2\\
  & \citet{park202fair} &  79.1$\pm$0.4 & 6.5$\pm$0.4\\
\hline
\label{table: eo_celeba_am}
\end{tabular}
\end{table}

\begin{table}[htbp]
\centering
\caption{Accuracy (\%) and Equalized Odds (EO) (\%) on CelebA with target attribute ``Attractive'' and sensitive attributes ``Young'' under different desired levels ($\delta$'s) of EO (\%) constraints. \textbf{Boldface} is used when MBS is better than \citet{park202fair} both in terms of accuracy and EO.}
 \begin{tabular}{c  c  c c   } 
\hline
 & $\delta$ & ACC & EO \\ 
 \hline
  \multirow{ 5}{*}{Validation} 
  & $\infty$ & 81.3$\pm$0.0 & 27.7$\pm$0.4\\
  & 10 & 80.2$\pm$0.1 & 10.0$\pm$0.0\\
  &5 & 79.2$\pm$0.1 & 4.9$\pm$0.1\\
  &1 & 78.2$\pm$0.2 & 1.0$\pm$0.0\\
  & \citet{hardt2016equality} &  70.4$\pm$0.1 & 0.4$\pm$0.2 \\
  \hline
  \multirow{ 6}{*}{Test}
  & $\infty$ & 82.0$\pm$0.0 & 24.7$\pm$0.8\\
  & 10 & \textbf{79.7$\pm$0.1} & \textbf{10.9$\pm$0.5}\\
  &5 & 78.2$\pm$0.0 & 6.6$\pm$0.8\\
  &1 & 76.9$\pm$0.0 & 2.0$\pm$0.7\\
  & \citet{hardt2016equality} &  71.4$\pm$0.3 & 3.4$\pm$0.7\\
  & \citet{park202fair} &  79.1$\pm$0.5 & 12.4$\pm$0.5\\
\hline
\label{table: eo_celeba_ay}
\end{tabular}
\end{table}

\begin{table}[htbp]
\centering
\caption{Accuracy (\%) and Equalized Odds (EO) (\%) on CelebA with target attribute ``Big\_Nose'' and sensitive attributes ``Male'' under different desired levels ($\delta$'s) of EO (\%) constraints. \textbf{Boldface} is used when MBS is better than \citet{park202fair} both in terms of accuracy and EO.}
 \begin{tabular}{c  c  c c   } 
\hline
 & $\delta$ & ACC & EO \\  
 \hline
  \multirow{ 5}{*}{Validation} 
  & $\infty$ & 82.9$\pm$0.6 & 43.3$\pm$2.1\\
  & 10 & 82.7$\pm$0.1 & 9.9$\pm$0.1\\
  &5 & 82.3$\pm$0.1 & 5.0$\pm$0.0\\
  &1 & 81.8$\pm$0.1 & 0.9$\pm$0.0\\
  & \citet{hardt2016equality} & 76.2$\pm$0.2  & 0.5$\pm$0.2 \\
  \hline
  \multirow{ 6}{*}{Test}
  & $\infty$ & 84.1$\pm$0.1 & 44.9$\pm$2.5\\
  & 10 & 83.5$\pm$0.2 & 8.4$\pm$1.2\\
  &5 & \textbf{83.3$\pm$0.2} & \textbf{4.3$\pm$1.0}\\
  &1 & 83.2 $\pm$0.1& 1.7$\pm$0.3\\
  & \citet{hardt2016equality} &  78.2$\pm$0.0 &  0.5$\pm$0.0 \\
  & \citet{park202fair} &  82.9$\pm$0.4 & 4.7$\pm$0.5\\
\hline
\label{table: eo_celeba_bm}
\end{tabular}
\end{table}

\begin{table}[htbp]
\centering
\caption{Accuracy (\%) and Equalized Odds (EO) (\%) on CelebA with target attribute ``Big\_Nose'' and sensitive attributes ``Young'' under different desired levels ($\delta$'s) of EO (\%) constraints. \textbf{Boldface} is used when MBS is better than \citet{park202fair} both in terms of accuracy and EO.}
 \begin{tabular}{c  c  c c   } 
\hline
 & $\delta$ & ACC & EO \\  
 \hline
  \multirow{ 5}{*}{Validation} 
  & $\infty$ & 83.5$\pm$0.0 & 21.5$\pm$0.1\\
  & 10 & 83.3$\pm$0.0 & 9.7$\pm$0.1\\
  &5 & 83.0$\pm$0.1 & 4.9$\pm$0.1\\
  &1 & 82.2$\pm$0.2 & 1.0$\pm$0.0\\
  & \citet{hardt2016equality} &  79.0$\pm$0.0 & 0.6$\pm$0.1\\
  \hline
  \multirow{ 6}{*}{Test}
  & $\infty$ & 84.5$\pm$0.1 & 21.0$\pm$0.1\\
  & 10 & 84.3$\pm$0.1 & 10.6$\pm$0.6\\
  &5 & \textbf{84.2$\pm$0.1} & \textbf{4.1$\pm$0.8}\\
  &1 & 83.6$\pm$0.2 & 2.4$\pm$0.6\\
  & \citet{hardt2016equality} &  80.1$\pm$0.1 & 1.3$\pm$0.3\\
  & \citet{park202fair} &  84.1$\pm$0.5 & 4.8$\pm$0.3\\
\hline
\label{table: eo_celeba_by}
\end{tabular}
\end{table}

\begin{table}[htbp]
\centering
\caption{Accuracy (\%) and Equalized Odds (EO) (\%) on CelebA with target attribute ``Bag\_Under\_Eyes'' and sensitive attributes ``Male'' under different desired levels ($\delta$'s) of EO (\%) constraints. \textbf{Boldface} is used when MBS is better than \citet{park202fair} both in terms of accuracy and EO.}
 \begin{tabular}{c  c  c c   } 
\hline
 & $\delta$ & ACC & EO \\  
 \hline
  \multirow{ 5}{*}{Validation} 
  & $\infty$ & 84.5$\pm$0.0 & 19.8$\pm$3.8\\
  & 10 & 84.4$\pm$0.0 & 8.3$\pm$0.7\\
  &5 & 84.2$\pm$0.1 & 5.0$\pm$0.0\\
  &1 & 82.3$\pm$0.1 & 1.0$\pm$0.0 \\
  & \citet{hardt2016equality} & 79.8$\pm$0.2  & 0.4$\pm$0.1 \\
  \hline
  \multirow{ 6}{*}{Test}
  & $\infty$ & 85.2$\pm$0.1 & 19.0$\pm$3.7\\
  & 10 & 85.0$\pm$0.0 & 8.2$\pm$0.7\\
  &5 & 84.7$\pm$0.1 & 5.6$\pm$0.2\\
  &1 & 83.1$\pm$0.0 & 1.7$\pm$0.2\\
  & \citet{hardt2016equality} &  80.2$\pm$0.2 & 1.7$\pm$0.4 \\
  & \citet{park202fair} &  83.4$\pm$0.6 & 3.0$\pm$0.4\\
\hline
\label{table: eo_celeba_em}
\end{tabular}
\end{table}

\begin{table}[htbp]
\centering
\caption{Accuracy (\%) and Equalized Odds (EO) (\%) on CelebA with target attribute ``Bag\_Under\_Eyes'' and sensitive attributes ``Young'' under different desired levels ($\delta$'s) of EO (\%) constraints. \textbf{Boldface} is used when MBS is better than \citet{park202fair} both in terms of accuracy and EO.}
 \begin{tabular}{c  c  c c   } 
\hline
 & $\delta$ & ACC & EO \\  
 \hline
  \multirow{ 5}{*}{Validation} 
  & $\infty$ & 84.6$\pm$0.1 & 18.2$\pm$1.1\\
  & 10 & 84.6$\pm$0.0 & 9.9$\pm$0.0\\
  &5 & 84.4$\pm$0.0 & 5.0$\pm$0.0\\
  &1 & 83.0$\pm$0.1 & 1.0$\pm$0.0 \\
  & \citet{hardt2016equality} & 81.1$\pm$0.2  & 0.5$\pm$0.2 \\
  \hline
  \multirow{ 6}{*}{Test}
  & $\infty$ & 85.6$\pm$0.3 & 13.7$\pm$1.9\\
  & 10 & 85.1$\pm$0.1 & 9.5$\pm$1.0\\
  &5 & 85.1$\pm$0.0 & 5.4$\pm$0.5\\
  &1 & 83.6$\pm$0.2 & 1.8$\pm$0.3\\
  & \citet{hardt2016equality} &  81.7$\pm$0.2 & 1.6$\pm$0.5 \\
  & \citet{park202fair} &  83.5$\pm$0.3 & 1.6$\pm$0.3\\
\hline
\label{table: eo_celeba_ey}
\end{tabular}
\end{table}

%% file: appendix/sensitivity_analysis.tex
\label{appendix: sensitivity_analysis}
\def\gv{{\boldsymbol{\gamma}}}

For the sake of convenience, we assume that the classes are valued as $\{-1, +1\}$ instead of $0, 1$. In such a case, the  unconstrained optimal classifier that has the form $ {Y}^{*}(X) = \sign( \hat{p}(Y=1|X) - 0.5) $ and constrained optimal classifiers have form $ {Y}_{\gv}^{*}(X) = {{Y}^{*}(X)} \sign\left( \sum_{j} \gamma_j s_j(X) - 1\right) $, i.e. ones that have highest accuracy subject to restriction $CC \leq \delta $.

{In the MBS implementation (Section~\ref{sec: method}),} we replace the unknown ground-truth probabilities $ p(Y|X) $ and $p(A|X)$ ($p(Y, A|X)$ for EO) with their estimations, which we denote as $ \hat{p}(Y|X) $ and $ \hat{p}(A|X)$ ($ \hat{p}(Y, A|X)$), respectively. 
The empirical version of the unconstrained classifier has the form $ \hat{Y}(X) = \sign( \hat{p}(Y=1|X) - 0.5) $ and we are looking for a classifier of the form $\check{Y}_{\gv}(X) = \hat{Y}(X) \sign\{ \sum_{j} \gamma_j \hat{s}_{j}(X) - 1\} $, i.e. our goal is to fit a set of $k$ parameters $\gv = (\gamma_j) \in \R^{k}$. Recall that the composite criterion has the form \EqRef{composite criteria}, and the score functions are defined in \EqRef{bias_score}, \EqRef{f_j_definition}.

We assume that we choose the score weights $\gv = (\gamma_1, \dots, \gamma_k)$ based on the observed validation sample $ \{ (X_i, Y_i, {A_{1i}, A_{2i}, \dots, A_{ki}})\}_{i = 1}^{N}$
of size $N$ by maximizing the corresponding empirical accuracy, while maintaining the restriction on empirical parity. Set,
\begin{align*}
    \widehat{Acc}(\check{Y}) &= \frac{1}{N} \sum_{i = 1}^{N} \Ind\{ Y_i = \check{Y}(X_i) \},
    \\
    \widehat{CC}(\check{Y}) &= \max_{j \leq k} | \widehat{C}_j(\check{Y})| \\
    \widehat{C}_{j}(\check{Y}) &= \frac{1}{N} \sum_{i = 1}^{N} \Ind\{ \check{Y}(X_i) = +1 \} \left[ \frac{\Ind\{A_{ji} = a_j\}}{\Pr(A_{ji} = a_j)} - \frac{\Ind\{A_{ji} = b_j\}}{\Pr(A_{ji} = b_j)}\right]\,.
\end{align*}

For a given level $\delta > 0$ we fit the parameters on validation as follows
\begin{equation}\label{validation_problem_statement}
    \hat{\gv} = \arg\max\{\widehat{Acc}(\check{Y}_\gv) : \widehat{DP}(\check{Y}_\gv) \leq \delta\}\,.
\end{equation}
The goal of this section is to demonstrate that, if we are provided an accurate enough estimate $\hat{p}$ and a sufficiently large validation set $N$, we can have guaranties for the empirically derived estimator $ \check{Y}_{\hat{\gv}}(X) $. We deliberately do not touch the question of what the estimation error 
 of $p(Y, A|X)$ could be, which may depend on various factors such as training sample size, input dimensionality, and complexity of the parametrized model \citep{vapnik1998statistical, boucheron2005theory}, as well as the training algorithm \citep{bousquet2002stability, hardt2016train, klochkov2021stability}. 

\begin{theorem}\label{theorem:sensitivity}
Assume the following set of conditions:
\begin{enumerate}
    \item We have estimation of conditional probabilities that satisfy moment bounds (for $p \geq 1$),
    \begin{align*}
        \E^{1/p} |\hat{p}(Y = 1|X) - p(Y = 1|X)|^{p} &\leq \eps_{p}, \\
        \E^{1/p} |\hat{p}(A_j = a_j|X) - p(A_j = b_j|X)|^{p} &\leq \eps_{p}, \qquad \forall j = 1, \dots, k, \\
        \E^{1/p} |\hat{p}(A_j = b_j|X) - p(A_j = b_j|X)|^{p} &\leq \eps_{p}, \qquad \forall j = 1, \dots, k;
    \end{align*}
    \item The distribution satisfies the \emph{margin assumption} $ \eta(X) > \eta^{*} $ for all $X$ (known as Mammen-Tsybakov condition \citep{mammen1999smooth});
    \item All groups have significant size: we have $ \Pr(A_j = a_j),\Pr(A_j = b_j) \geq p_0 $;
    \item Score distribution regularity: the distribution of scores $\sv(X) = (s_1(X), \dots, s_k(X))$ is supported on a ball $\{ s \in \R^{k}: \| s \| \leq B \}$; Moreover, for every $\|\gv\| = 1$, the density of $\gv^{\T} \sv(X)$ is bounded by a constant $ L $;
    \item The weights corresponding to optimal $Y^{*}_{\gv}$ under $CC(Y^{*}_{\gv})\leq \delta / 2$ satisfy $\|\gv\|_{1} \leq R $;
\end{enumerate}
There is a constant $C = C(\eta^*, p_0, L, R, B, \delta) $, such that, for any $r \in (0, 1)$ that satisfies,
\begin{equation}\label{reviewer4_condition}
    C \left( \sqrt{\frac{k \log(kN / r)}{N}} + (\sqrt{k}\eps_p)^{1 - \frac{1}{1 + p}} \right) \leq 1,
\end{equation}
we have that, with probability at least $1 - r$,
\begin{equation}\label{acc_dp_bounds}
\begin{aligned}
    Acc(\check{Y}_{\hat{\gv}}) &\geq Acc^{*}(\delta) - C \left((\sqrt{k}\eps_p)^{1 - \frac{1}{p + 1}} + \sqrt{\frac{k\log (kN/r)}{N}} \right) , \\
    CC(\check{Y}_{\hat{\gv}}) &\leq \delta  + C \sqrt{\frac{k \log (kN/r)}{N}} ,
\end{aligned}
\end{equation}
where $  Acc^{*}(\delta) $ is the accuracy of the optimal classifier under the constraints $ CC \leq \delta$.
\end{theorem}

Before we move on to the proof, let us briefly comment on each condition. The first condition requires that we have a good estimation of probabilities in the form of moment bound. We note that bounds of this form often appear in theoretical analysis of non-parametric regression \cite{Tsybakov2009Nonparametric}, including ones based on neural networks \cite{hu2021regularization}. Although they are standard in non-parametric regression, there is no way to verify them in practice without making structural assumption about the underlying distribution. 
The margin condition is very popular in Statistical Learning Theory, and in our particular case, it allows controlling the accuracy of the estimator under perturbations of the conditionals. {We note that \cite{denis2021fairness} also assume a form of margin condition for the multi-class fair classification.} The third condition is somewhat realistic and simply makes sure we have an appropriate criterion evaluation based on the validation sample. The fourth condition is a stronger version of the condition in Theorem~\ref{main_thm}, where we require that the distribution of $\gv^{\T} \sv(X)$ is continuous. We prefer to have a much stronger version to avoid technical difficulties in the proof. As for the fifth condition, it is pretty much guaranteed that the vectors $\gv_{\delta}$ are bounded, which we show for instance as part of the proof (see \EqRef{z_f_equation_} and the line below it). However, it is hard to precisely characterize the value of $R$ based on characteristic of this distribution, so we prefer to state it as a condition.

Furthermore we note that practical implementations based on Bayes optimal classifiers are often based on closed form derivations of the optimal weights \cite{denis2021fairness, zeng2022bayes}. We do not provide these, instead our method treats the bias scores as features when fitting the modification rule. As a result, the error of estimation of the conditionals does not propagate as much into the fairness constraint. See also some empirical studies in Appendix~\ref{appendix: ablation}.

\begin{proof}
We compare the accuracy and composite criterion of $Y^*_\gv$ and $\check{Y}_{\hat{\gv}}$ in three steps:
\begin{enumerate}
    \item \textit{Approximation on population level}: we bound the discrepancy in the population statistics ($Acc$ and $CC$) of estimators $Y^*_{\gv}$ and $\check{Y}_{\gv}$, i.e. how the difference between conditionals $p(Y|X), p(A_j|X)$ and its estimation propagates into the accuracy and composite criterion.
    \item \textit{Approximation on validation set}: we then compare the population statistics and validation statistics ($\widehat{Acc}$, $\widehat{CC}$) simultaneously for all estimators $Y^{*}_{\gv}, \check{Y}_{\gv}$.
    \item \textit{Connecting validation and population}: once the first two steps are done, we use the standard trick in empirical process theory to connect the classifiers that are optimal in population and in validation.
\end{enumerate}
\textbf{Step 1 (approximation on population level).} We first check how the population accuracy and DP compare between the optimal $Y^{*}(X)$ and the one based on the estimated probabilities $ \hat{Y}(X) $.

For the unconstrained classifiers, thanks to the margin condition, we have by Markov inequality:
\begin{align*}
    \Pr(\hat{Y} \neq Y^*) &= \Pr( \hat{p}(Y|X) < 0.5 < p(Y|X) \;\text{ or }\; {p}(Y|X) < 0.5 < \hat{p}(Y|X)) \\
    & \leq \Pr( |\hat{p}(Y|X) - p(Y|X)| > \eta^*) \\
    & \leq \left(\frac{\eps_p}{\eta^*} \right)^{p}.
\end{align*}

In particular, this implies that $ Acc(\hat{Y}) \geq Acc(Y^*) - \left({\eps_p}/{\eta^*} \right)^{p}$. In order to compare the accuracies of constrained classifiers, we need to further compare the modification rules. Denote $\kappa_{\gv}(X) = \sign\{ \gv^{\T} \sv(X) - 1 \} $ for the ground truth score, and consider empirically derived rule $ \hat{\kappa}_{\gv}(X) = \sign\{ \gv^{\T} \hat{\sv}(X) - 1 \}  $,
where $ \hat{\sv}(X) $ are derived by replacing $p(A_j=a_j|X)$ and $p(A_j = b_j|X)$ with $ \hat{p}(A_j = a_j|X) $ and $ \hat{p}(A_j = b_j|X) $, respectively.
Let also $ \hat{f}_j(X)$, $\hat{\eta}(X)$ be the corresponding substitutes for $f_j(X), \eta(X)$. We have,
\begin{align*}
    \Pr(\kappa_\gv < \hat{\kappa}_{\gv}) & = \Pr(\gv^{\T} \sv(X) < 1 < \gv^{\T} \hat{\sv}(X)) \nonumber \\
    & = \Pr\left(\gv^{\T} \sv(X) \in \left[1 + \gv^{\T} ({\sv}(X) - \hat{\sv}(X))  ; 1 \right] \right) \nonumber \\
    & = \Pr\left(\tilde{\gv}^{\T} \sv(X) \in \left[1/\| \gv\| + \tilde{\gv}^{\T} ({\sv}(X) - \hat{\sv}(X))  ; 1 / \| \gv\| \right] \right), 
\end{align*}
where we denote the normalized weights $\tilde{\gv} = \gv / \| \gv\| $. Taking the union with the opposite inequality, we get that
\begin{equation}
    \Pr(\kappa_\gv \neq \hat{\kappa}_{\gv}) \leq \Pr\left(\tilde{\gv}^{\T} \sv(X) \in \left[1/\| \gv\| - |\tilde{\gv}^{\T} ({\sv}(X) - \hat{\sv}(X))|  ; 1 / \| \gv\| + |\tilde{\gv}^{\T} ({\sv}(X) - \hat{\sv}(X))|\right] \right) \,.\label{eq_gamma_s_in_interval}
\end{equation}
To proceed, we want to bound the value $\tilde{\gv}^{\T} ({\sv}(X) - \hat{\sv}(X))$.
Firstly, it is straightforward to see that $\E^{1/p} | \eta - \hat{\eta} |^p \leq 2\eps_{p}$. For the $f$-score, we see that
\[
    | f_j - \hat{f}_j | \leq \frac{1}{p_0} \Ind\{ \hat{Y} \neq Y^*\} + \frac{1}{p_0} | p(A_j = a_j|X) - 
    \hat{p}(A_j=a_j|X)| +\frac{1}{p_0} | p(A_j = b_j|X) - 
    \hat{p}(A_j=b_j|X)|,
\]
so that by the triangle inequality,
\[
    \E^{1/p} | f - \hat{f} |^p \leq p_0^{-1} \left(\frac{\eps_p}{\eta^*} \right) + 2 p_0^{-1} \eps_p \leq \frac{3 \eps_p}{p_0\eta^*}.
\]
Then, we can write
\begin{align*}
    \E^{1/p} | \hat{\eta} \tilde{\gv}^{\T} (s(X) - \hat{s}(X)) |^{p} & \leq \E^{1/p} \left| \frac{\hat{\eta}}{\eta} \sum_{j} (f_j - \hat{f}_j) \tilde{\gamma}_j\right|^{p} + \E^{1/p}\left| \left(\frac{\hat{\eta}}{\eta } - 1\right) \sum_j \tilde{\gamma}_j \hat{f}_j\right|^{p} \\
    & \leq \frac{1}{\eta^*} \E^{1/p} \left| \sum_{j} |f_j - \hat{f}_j|^{2}\right|^{p/2} + \frac{2\sqrt{k}}{p_0\eta^*} \E^{1/p} |\eta - \hat{\eta}|^{p} \\
    & \leq \frac{3\sqrt{k}}{p_0\eta^*} \eps_p\,.
\end{align*}
Using the Markov inequality, this gives us the bound for any $r \in (0, 1)$,
\[
    \Pr\left( |\hat{\eta} \tilde{\gv}^{\T} (s(X) - \hat{s}(X))| \leq r^{-1} \frac{5\sqrt{k}}{p_0\eta^*} \eps_p \right) \geq 1 - r^{p}\,.
\]
Using the moment bound on $\eta - \hat{\eta}$, we can also lowerbound $\hat{\eta}$,
\[
    \Pr(\hat{\eta} \geq \eta^{*} / 2) \leq \Pr(|\hat{\eta} - \eta| \leq \eta^{*} / 2) \leq 1 - \left(\frac{4\eps_p}{\eta^*}\right)^{p}\,.
\]
Taking the union bound of the two last displays, we get that
\[
    \Pr\left(|\tilde{\gv}^{\T} (s(X) - \hat{s}(X))| \leq r^{-1} \frac{6\sqrt{k}}{p_0(\eta^*)^{2}} \eps_p\right) \leq 1 - \left(\frac{4\eps_p}{\eta^*}\right)^{p} - r^{p}\, .
\]
Now, we can plug this back into \EqRef{eq_gamma_s_in_interval}. We get that,
\[
    \Pr(\kappa_{\gv} \neq \hat{\kappa}_{\gv}) \leq 2L r^{-1} \frac{6\sqrt{k}}{p_0(\eta^*)^{2}} \eps_p + r^{p} + \left(\frac{4\eps_p}{\eta^*}\right)^{p} \,.
\]
Optimizing $r = (12L\sqrt{k} \eps_p / (p p_0 (\eta^*)^2))^{\frac{1}{p + 1}} $, we get that
\[
    \Pr(\kappa_{\gv} \neq \hat{\kappa}_{\gv}) \leq 24 \left(\frac{L\sqrt{k} \eps_p}{p_0(\eta^*)^{2}}\right)^{1 - 1/(p + 1)} + \left(\frac{4\eps_p}{\eta^*}\right)^{p} \,.
\]
We now can derive using the triangle inequality,
\begin{align}
    |Acc(\check{Y}_\gv) - Acc(Y^{*}_{\gv})| & \leq \Pr(\check{Y}_\gv \neq Y^{*}_{\gv})\nonumber \\
    & \leq \Pr(\hat{Y} \neq Y^*) + \Pr(\kappa_\gv \neq \hat{\kappa}_{\gv})\nonumber \\
    & \leq 2 \left(\frac{4\eps_p}{\eta^*} \right)^{p} + 24 \left(\frac{L\sqrt{k}\eps_p}{p_0(\eta^*)^{2}}\right)^{1 - 1/(p + 1)}\nonumber \\
    & \leq L_1 (\sqrt{k}\eps_p)^{1 - 1/(p + 1)} \label{bound_on_accuracy},
\end{align}
where we assume that $ \eps_p < \eta^* / 4$ (condition \EqRef{reviewer4_condition}) and set $L_1 = 4/ \eta^{*} + 24 \max((L/(p_0\eta^*))^{1 - 1/(p + 1)}, 1) $.
Similarly,
\[
    |CC(\check{Y}_\gv) - CC(Y^{*}_{\gv})|  \leq \max_{j} |C_j(\check{Y}_\gv) - C_j({Y}^{*}_\gv)| \leq  p_{0}^{-1} \Pr(\check{Y}_\gv \neq Y^{*}_{\gv}) \leq L_2 (\sqrt{k}\eps_p)^{1 - 1/(p + 1)},
\]
where $L_2 = p_0^{-1} L_1$.

\textbf{Step 2 (approximation on validation set).} The next step is to derive uniform bounds for the empirical accuracy and CC. We have,
\begin{align*}
    \widehat{Acc}(\check{Y}_\gv) &= \frac{1}{2N} \sum_{i = 1}^{N} \hat{Y}(X_i) Y_{i} \sign(\gv^{\T}\hat{\sv}(X_i) - 1) + \frac{1}{2} \\
     \widehat{Acc}({Y}^{*}_\gv) &= \frac{1}{2N} \sum_{i = 1}^{N} {Y}^{*}(X_i) Y_{i} \sign(\gv^{\T}{\sv}(X_i) - 1) + \frac{1}{2}
\end{align*}
and notice that $ \E_{val} \widehat{Acc}(\check{Y}_\gv) = {Acc}(\check{Y}_\gv)$ and $\E_{val} \widehat{Acc}({Y}^{*}_\gv) = {Acc}({Y}^{*}_\gv)$, where we conventionally denote $ \E_{val} $ as the expectation (and below $\Pr_{val}$ for probability) w.r.t. the validation sample.

\begin{lemma}[\cite{vapnik1998statistical}, Theorem~5.3]
Suppose, we have a bounded function $ f(Z) \in [-1, 1] $, and $k$ arbitrary functions $g_1(X), g_2(X), \dots, g_k(X)$, and consider a class of functions $\{ \lambda_{\gv}(X) = f(Z) \sign\{\sum_{j} \gamma_j g_j(Z) - 1\}: \gv \in \R^k\}$.
Let $ X_1, \dots, X_N$ be i.i.d. Then, we have with probability $1 - \delta$ that for each $\gv$,
\[
    \left| \frac{1}{N} \sum_{i} \lambda_{\gv}(Z_i) - \E \lambda_{\gv}(Z) \right| \leq L_0 \sqrt{\frac{k \log (N / \delta)}{N} }\,.
\]
\end{lemma}
\begin{proof}
We simply need to observe that the shatter coefficient of set $ \{ \lambda_{\gv}(X) < \tau\} $ is bounded by $ (N + 1)^{k+1} $, then apply Theorem 5.3 from \cite{vapnik1998statistical}.
\end{proof}

For $Z=(X, Y)$ set $ f(Z) = \hat{Y}(X) Y / 2 $, $g_j(Z) = \hat{s}_j(X)$ to control the accuracy of $\check{Y}_{\gv}(X)$ and $ f(Z) = {Y}^*(X) Y / 2 $, $g_j(Z) = s_j(X)$, to control the accuracy of ${Y}_{\gv}^{*}(X)$. Furthermore, for $Z=(X, Y, A_{1}, \dots, A_{k})$ we can set $ f(Z) = \hat{Y}(X) \left[  \Pr(A_j=a_j)^{-1} \mathbf{1}\{A_j = a_j\} - \Pr(A_j=b_j)^{-1} \mathbf{1}\{A_j = b_j\} )\right] / 2 $, $g(Z) = \sign(\gv^{\T}\hat{\sv}(X) - 1)$ to get concentration of disparity $C_j$ for $\check{Y}_{\gv}(X)$, and similarly for ${Y}_{\gv}^{*}(X)$. Overall, we have that for any $r \in (0, 1)$ each of the bounds holds
\begin{align*}
    \left| \widehat{Acc}(\check{Y}_\gv) - {Acc}(\check{Y}_\gv) \right| &\leq  L_0 \sqrt{\frac{k\log(N/r)}{N}}  \qquad \text{w. p.} \qquad \geq 1 - r \\
    \left| \widehat{Acc}({Y}^{*}_\gv) - {Acc}({Y}^{*}_\gv) \right| &\leq L_0 \sqrt{\frac{k\log(N/r)}{N}}  \qquad \text{w. p.} \qquad \geq 1 - r\\
    \left| \widehat{C}_j(\check{Y}_\gv) - {C}_j(\check{Y}_\gv) \right| &\leq L_0 p_0^{-1} \sqrt{\frac{k\log(N/r)}{N}} \qquad \text{w. p.} \qquad \geq 1 - r \\
    \left| \widehat{C}_j({Y}^{*}_\gv) - {C}_j({Y}^{*}_\gv) \right| &\leq L_0 p_0^{-1} \sqrt{\frac{k\log(N/r)}{N}}  \qquad \text{w. p.} \qquad \geq 1 - r\\
\end{align*}
If we take $r = r / k$ for the bound on each $C_j$ and then take a union bound, we obtain a bound on the difference in the composite criterion. Overall, we have that with probability at least $1 - r$,
\begin{align*}
\left| \widehat{Acc}(\check{Y}_\gv) - {Acc}(\check{Y}_\gv) \right| &\leq  L_3 \sqrt{\frac{k\log(N/r)}{N}} \\
    \left| \widehat{Acc}({Y}^{*}_\gv) - {Acc}({Y}^{*}_\gv) \right| &\leq L_{3}\sqrt{\frac{k\log(N/r)}{N}} \\
    \left| \widehat{CC}(\check{Y}_\gv) - {CC}(\check{Y}_\gv) \right| &\leq L_3 \sqrt{\frac{k\log(kN/r)}{N}}  \\
    \left| \widehat{CC}({Y}^{*}_\gv) - {CC}({Y}^{*}_\gv) \right| &\leq L_3 \sqrt{\frac{k\log(kN/r)}{N}}\,.
\end{align*}
where we set $L_3 = L_0 p_{0}^{-1} \sqrt{\log 4}$.

\paragraph{Step 3 (connecting validation and population).} Let $\gv$ corresponds to optimal $ Y^{*}_{\gv}$ under $CC({Y}^{*}_{\gv}) \leq \delta$ and denote its accuracy $Acc^*(\delta)$. Then, let $\gv'$ corresponds to the the optimal $ Acc(Y^{*}_{\gv'}) $ under the constraint $ CC({Y}^{*}_{\gv'}) \leq \delta - \epsilon $. At the end of this section we show that 
\begin{equation}\label{acc_delta_minus_eps_2}
    Acc(Y^{*}_{\gv'}) = Acc^*(\delta - \epsilon) \geq Acc^*(\delta) - \epsilon R \;\;\; \text{for} \;\;\; \epsilon < \delta / 2,
\end{equation}
and let us assume for now that it is true. Set $ \epsilon_1 = L_3 \sqrt{\frac{k\log(kN/r)}{N}}, \epsilon_2 = L_2 \eps_p ^{1 - 1/(p + 1)}, \epsilon = \epsilon_1 + \epsilon_2$, and note that by assumption \EqRef{reviewer4_condition}, $ \epsilon < \delta / 2$. Then, $\widehat{CC}(\check{Y}_{\gv'}) \leq  {CC}(\check{Y}_{\gv'}) + L_3 \sqrt{\frac{k\log(kN/\gamma)}{N}} \leq {CC}(\check{Y}^{*}_{\gv'}) + \epsilon = \delta$, which means that the problem \EqRef{validation_problem_statement} is feasible, and also  $\widehat{Acc}(\check{Y}_{\hat{\gv}}) \geq \widehat{Acc}(\check{Y}_{\gv'}) \geq {Acc}(\check{Y}_{\gv'}) - \epsilon_1 \overset{\EqRef{bound_on_accuracy}}\geq Acc(Y^{*}_{\gv'}) - \epsilon_2 - \epsilon_1 = Acc^*(\delta - \epsilon) - \epsilon$. Finally, it implies that
\begin{align*}
    {Acc}(\check{Y}_{\hat{\gv}}) & \geq Acc^*(\delta) - (2 + R) \epsilon \\
    CC(\check{Y}_{\hat{\gv}}) & \leq \delta - \epsilon_1.
\end{align*}
Substituting $\epsilon, \epsilon_1$, we finally get the bound stated in \EqRef{acc_dp_bounds}.

\paragraph{Check of \EqRef{acc_delta_minus_eps_2}.} Let us denote $ \gv_{\delta} $ to be (some) optimal set of weights, corresponding to the classifier that maximizes the accuracy under restriction $CC(Y^*_{\gv}) \leq \delta$. We show that for $\epsilon < \delta$, 
\[
    Acc(\delta - \epsilon) - Acc(\delta) \leq \epsilon \| \gv_{\delta - \epsilon}\|_1,
\]
which thanks to condition 5. yields \EqRef{acc_delta_minus_eps_2}. To show that, we need to recall the proof of Theorem~\ref{main_thm}, where $\gv_{\delta}$ comes from the dual LP problem (Lemma~\ref{lemma_linear_flipping_rule}), and the value $Acc^*(\delta)$ corresponds to the maximal objective in the primal and minimal objective in the dual LP. Changing $\delta \mapsto \delta - \epsilon$ simply means that we change the vector $b$ in \EqRef{dual_problem} to a new $b^{new} = b - \eps \begin{pmatrix} \mathbf{1}_{k} \\ -\mathbf{1}_{k} \end{pmatrix}$, where $\mathbf{1}_k$ denotes a column with $k$ ones. Let $z^{new}, \lambda^{new}$ denotes the solution to dual LP with $b$ replaced by $b^{new}$ in \EqRef{dual_problem}. Then, since in the dual LP the constraints do not depend on $b$, the value at $z^{new}, \lambda^{new}$ serves as an upperbound for the old problem. Therefore, $Acc(\delta) - Acc(\delta - \eps) \leq z^{new} (b^{new} - b) \leq  \eps \|\gv_{\delta - \eps}\|_1$, since the corresponding $\gv_{\delta - \eps}$ has the coordinates $ z_{j} - z_{j + k} $.
\end{proof}

%% file: appendix/reduction.tex
Here, we include Reduction method \citep{agarwal2018reductions} as an additional baseline, and we use the same network architecture and hyperparameters as for the other methods.
\begin{figure}[htbp]
    \centering
    \begin{subfigure}{0.45\textwidth}
        \includegraphics[width=\textwidth]{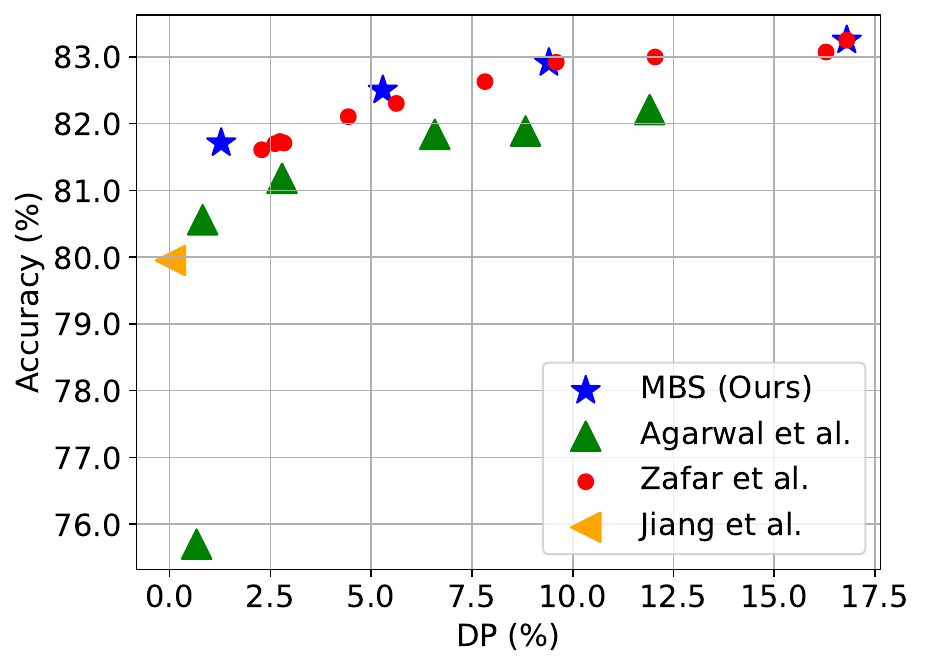}
        \caption{}\label{fig_dp_adult}
    \end{subfigure}
    \centering
    \begin{subfigure}{0.45\textwidth}
        \includegraphics[width=\textwidth]{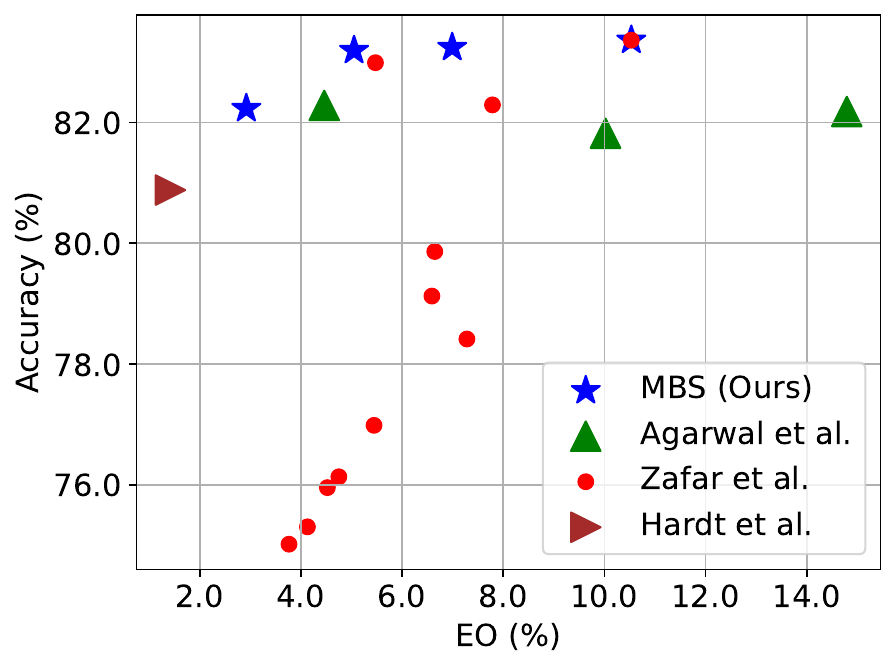}
        \caption{}\label{fig_eo_adult}
    \end{subfigure}
    \caption{Accuracy (\%) vs (a) Demographic Parity (DP) (\%) and (b) Equalized Odds (EO) trade-offs on Adult Census. Desired $\delta=\infty\text{ (unconstrained), }10\%\text{, } 5\%\text{, } \text{ and } 1\%$.}
    \label{fig: eo_dp_results}
\end{figure}

%% file: main.bbl
\begin{thebibliography}{37}
\providecommand{\natexlab}[1]{#1}
\providecommand{\url}[1]{\texttt{#1}}
\expandafter\ifx\csname urlstyle\endcsname\relax
  \providecommand{\doi}[1]{doi: #1}\else
  \providecommand{\doi}{doi: \begingroup \urlstyle{rm}\Url}\fi

\bibitem[Agarwal et~al.(2018)Agarwal, Beygelzimer, Dud{\'\i}k, Langford, and Wallach]{agarwal2018reductions}
Alekh Agarwal, Alina Beygelzimer, Miroslav Dud{\'\i}k, John Langford, and Hanna Wallach.
\newblock A reductions approach to fair classification.
\newblock In \emph{International conference on machine learning}, pp.\  60--69. PMLR, 2018.

\bibitem[Angwin et~al.(2015)Angwin, Larson, Mattu, and Kirchner]{angwin2015machine}
J.~Angwin, J.~Larson, S.~Mattu, and L.~Kirchner.
\newblock {Machine bias: There’s software used across the country to 272 predict future criminals. And it’s biased against blacks}.
\newblock 2015.
\newblock URL \url{https://www.propublica.org/article/machine-bias-risk-assessments-incriminal-sentencing}.

\bibitem[Bhattacharya et~al.(2017)Bhattacharya, Kanaya, and Stevens]{bhattacharya2017university}
Debopam Bhattacharya, Shin Kanaya, and Margaret Stevens.
\newblock Are university admissions academically fair?
\newblock \emph{Review of Economics and Statistics}, 99\penalty0 (3):\penalty0 449--464, 2017.

\bibitem[Boucheron et~al.(2005)Boucheron, Bousquet, and Lugosi]{boucheron2005theory}
St{\'e}phane Boucheron, Olivier Bousquet, and G{\'a}bor Lugosi.
\newblock Theory of classification: A survey of some recent advances.
\newblock \emph{ESAIM: probability and statistics}, 9:\penalty0 323--375, 2005.

\bibitem[Bousquet \& Elisseeff(2002)Bousquet and Elisseeff]{bousquet2002stability}
Olivier Bousquet and Andr{\'e} Elisseeff.
\newblock Stability and generalization.
\newblock \emph{The Journal of Machine Learning Research}, 2:\penalty0 499--526, 2002.

\bibitem[Caton \& Haas(2023)Caton and Haas]{caton2023fairness}
Simon Caton and Christian Haas.
\newblock Fairness in machine learning: A survey.
\newblock \emph{ACM Computing Surveys}, 2023.

\bibitem[Chouldechova(2017)]{chouldechova2017fair}
Alexandra Chouldechova.
\newblock Fair prediction with disparate impact: A study of bias in recidivism prediction instruments.
\newblock \emph{Big data}, 5\penalty0 (2):\penalty0 153--163, 2017.

\bibitem[Cotter et~al.(2019)Cotter, Jiang, Gupta, Wang, Narayan, You, and Sridharan]{cotter2019optimization}
Andrew Cotter, Heinrich Jiang, Maya~R Gupta, Serena~Lutong Wang, Taman Narayan, Seungil You, and Karthik Sridharan.
\newblock Optimization with non-differentiable constraints with applications to fairness, recall, churn, and other goals.
\newblock \emph{J. Mach. Learn. Res.}, 20\penalty0 (172):\penalty0 1--59, 2019.

\bibitem[Denis et~al.(2021)Denis, Elie, Hebiri, and Hu]{denis2021fairness}
Christophe Denis, Romuald Elie, Mohamed Hebiri, and Fran{\c{c}}ois Hu.
\newblock Fairness guarantee in multi-class classification.
\newblock \emph{arXiv preprint arXiv:2109.13642}, 2021.

\bibitem[Donini et~al.(2018)Donini, Oneto, Ben-David, Shawe-Taylor, and Pontil]{donini2018empirical}
Michele Donini, Luca Oneto, Shai Ben-David, John~S Shawe-Taylor, and Massimiliano Pontil.
\newblock Empirical risk minimization under fairness constraints.
\newblock \emph{Advances in neural information processing systems}, 31, 2018.

\bibitem[Gaucher et~al.(2023)Gaucher, Schreuder, and Chzhen]{gaucher2023fair}
Solenne Gaucher, Nicolas Schreuder, and Evgenii Chzhen.
\newblock Fair learning with wasserstein barycenters for non-decomposable performance measures.
\newblock In \emph{International Conference on Artificial Intelligence and Statistics}, pp.\  2436--2459. PMLR, 2023.

\bibitem[Hardt et~al.(2016{\natexlab{a}})Hardt, Price, and Srebro]{hardt2016equality}
Moritz Hardt, Eric Price, and Nathan Srebro.
\newblock Equality of opportunity in supervised learning.
\newblock In \emph{The Conference on Neural Information Processing Systems (NeurIPS)}, 2016{\natexlab{a}}.

\bibitem[Hardt et~al.(2016{\natexlab{b}})Hardt, Recht, and Singer]{hardt2016train}
Moritz Hardt, Ben Recht, and Yoram Singer.
\newblock Train faster, generalize better: Stability of stochastic gradient descent.
\newblock In \emph{International conference on machine learning}, pp.\  1225--1234. PMLR, 2016{\natexlab{b}}.

\bibitem[He et~al.(2016)He, Zhang, Ren, and Sun]{he2016deep}
Kaiming He, Xiangyu Zhang, Shaoqing Ren, and Jian Sun.
\newblock Deep residual learning for image recognition.
\newblock In \emph{IEEE conference on computer vision and pattern recognition (CVPR)}, 2016.

\bibitem[Hu et~al.(2021)Hu, Wang, Lin, and Cheng]{hu2021regularization}
Tianyang Hu, Wenjia Wang, Cong Lin, and Guang Cheng.
\newblock Regularization matters: A nonparametric perspective on overparametrized neural network.
\newblock In \emph{International Conference on Artificial Intelligence and Statistics}, pp.\  829--837. PMLR, 2021.

\bibitem[Jang et~al.(2022)Jang, Shi, and Wang]{jang2022group}
Taeuk Jang, Pengyi Shi, and Xiaoqian Wang.
\newblock Group-aware threshold adaptation for fair classification.
\newblock In \emph{Proceedings of the AAAI Conference on Artificial Intelligence}, volume~36, pp.\  6988--6995, 2022.

\bibitem[Jiang et~al.(2019)Jiang, Pacchiano, Stepleton, Jiang, and Chiappa]{jiang2019wasserstein}
Ray Jiang, Aldo Pacchiano, Tom Stepleton, Heinrich Jiang, and Silvia Chiappa.
\newblock Wasserstein fair classification.
\newblock In \emph{Conference on Uncertainty in Artificial Intelligence}, 2019.

\bibitem[Klochkov \& Zhivotovskiy(2021)Klochkov and Zhivotovskiy]{klochkov2021stability}
Yegor Klochkov and Nikita Zhivotovskiy.
\newblock {Stability and Deviation Optimal Risk Bounds with Convergence Rate $ O (1/n) $}.
\newblock \emph{Advances in Neural Information Processing Systems}, 34:\penalty0 5065--5076, 2021.

\bibitem[Kohavi(1996)]{kohavi1996scaling}
Ron Kohavi.
\newblock Scaling up the accuracy of naive-bayes classifiers: A decision-tree hybrid.
\newblock \emph{Knowledge Discovery and Data Mining}, 1996.

\bibitem[Liu et~al.(2015)Liu, Luo, Wang, and Tang]{liu2015deep}
Ziwei Liu, Ping Luo, Xiaogang Wang, and Xiaoou Tang.
\newblock Deep learning face attributes in the wild.
\newblock In \emph{International Conference on Computer Vision (ICCV)}, 2015.

\bibitem[Lundberg \& Lee(2017)Lundberg and Lee]{lundburg2017unified}
S.~M. Lundberg and S.~I. Lee.
\newblock A unified approach to interpreting model predictions.
\newblock \emph{Advances in neural information processing systems}, 2017.

\bibitem[Mammen \& Tsybakov(1999)Mammen and Tsybakov]{mammen1999smooth}
Enno Mammen and Alexandre~B Tsybakov.
\newblock Smooth discrimination analysis.
\newblock \emph{The Annals of Statistics}, 27\penalty0 (6):\penalty0 1808--1829, 1999.

\bibitem[Menon \& Williamson(2018)Menon and Williamson]{menon2018cost}
Aditya~Krishna Menon and Robert~C Williamson.
\newblock The cost of fairness in binary classification.
\newblock In \emph{Conference on Fairness, accountability and transparency}, pp.\  107--118. PMLR, 2018.

\bibitem[Park et~al.(2022)Park, Lee, Lee, Hwang, Kim, and Byun]{park202fair}
Sungho Park, Jewook Lee, Pilhyeon Lee, Sunhee Hwang, Dohyung Kim, and Hyeran Byun.
\newblock Fair contrastive learning for facial attribute classification.
\newblock In \emph{The IEEE/CVF Conference on Computer Vision and Pattern Recognition (CVPR)}, 2022.

\bibitem[Radford et~al.(2021)Radford, Kim, Hallacy, Ramesh, Goh, Agarwal, Sastry, Askell, Mishkin, Clark, et~al.]{radford2021learning}
Alec Radford, Jong~Wook Kim, Chris Hallacy, Aditya Ramesh, Gabriel Goh, Sandhini Agarwal, Girish Sastry, Amanda Askell, Pamela Mishkin, Jack Clark, et~al.
\newblock Learning transferable visual models from natural language supervision.
\newblock In \emph{International conference on machine learning}, pp.\  8748--8763. PMLR, 2021.

\bibitem[Ramaswamy et~al.(2021)Ramaswamy, Kim, and Russakovsky]{ramaswamy2021fair}
Vikram~V Ramaswamy, Sunnie~SY Kim, and Olga Russakovsky.
\newblock Fair attribute classification through latent space de-biasing.
\newblock In \emph{Proceedings of the IEEE/CVF conference on computer vision and pattern recognition}, pp.\  9301--9310, 2021.

\bibitem[Rezaei et~al.(2021)Rezaei, Liu, Memarrast, and Ziebart]{rezaei2021robust}
Ashkan Rezaei, Anqi Liu, Omid Memarrast, and Brian~D Ziebart.
\newblock Robust fairness under covariate shift.
\newblock In \emph{Proceedings of the AAAI Conference on Artificial Intelligence}, volume~35, pp.\  9419--9427, 2021.

\bibitem[Tsybakov(2009)]{Tsybakov2009Nonparametric}
Alexandre~B. Tsybakov.
\newblock \emph{{Introduction to Nonparametric Estimation}}.
\newblock Springer, 2009.
\newblock ISBN 978-0-387-79051-0.

\bibitem[Vapnik(1998)]{vapnik1998statistical}
Vladimir Vapnik.
\newblock {Statistical learning theory. John Wiley \& Sons, Chichester}.
\newblock 1998.

\bibitem[Wang et~al.(2021)Wang, Liu, and Levy]{wang2021fair}
Jialu Wang, Yang Liu, and Caleb Levy.
\newblock Fair classification with group-dependent label noise.
\newblock In \emph{Proceedings of the 2021 ACM conference on fairness, accountability, and transparency}, pp.\  526--536, 2021.

\bibitem[Wang et~al.(2022)Wang, Wang, and Liu]{wang2022understanding}
Jialu Wang, Xin~Eric Wang, and Yang Liu.
\newblock Understanding instance-level impact of fairness constraints.
\newblock In \emph{International Conference on Machine Learning}, pp.\  23114--23130. PMLR, 2022.

\bibitem[Xian et~al.(2023)Xian, Yin, and Zhao]{xian2023fair}
Ruicheng Xian, Lang Yin, and Han Zhao.
\newblock Fair and optimal classification via post-processing.
\newblock In \emph{International Conference on Machine Learning}, pp.\  37977--38012. PMLR, 2023.

\bibitem[Yao \& Liu(2023)Yao and Liu]{yao2023understanding}
Yuanshun Yao and Yang Liu.
\newblock Understanding unfairness via training concept influence.
\newblock \emph{arXiv preprint arXiv:2306.17828}, 2023.

\bibitem[Zafar et~al.(2017)Zafar, Valera, Rogriguez, and Gummadi]{zafar2017fairness}
Muhammad~Bilal Zafar, Isabel Valera, Manuel~Gomez Rogriguez, and Krishna~P Gummadi.
\newblock Fairness constraints: Mechanisms for fair classification.
\newblock In \emph{Artificial intelligence and statistics}, pp.\  962--970. PMLR, 2017.

\bibitem[Zafar et~al.(2019)Zafar, Valera, Gomez-Rodriguez, and Gummadi]{zafar2019fairness}
Muhammad~Bilal Zafar, Isabel Valera, Manuel Gomez-Rodriguez, and Krishna~P Gummadi.
\newblock Fairness constraints: A flexible approach for fair classification.
\newblock \emph{The Journal of Machine Learning Research}, 20\penalty0 (1):\penalty0 2737--2778, 2019.

\bibitem[Zeng et~al.(2022)Zeng, Dobriban, and Cheng]{zeng2022bayes}
Xianli Zeng, Edgar Dobriban, and Guang Cheng.
\newblock Bayes-optimal classifiers under group fairness.
\newblock \emph{arXiv preprint arXiv:2202.09724}, 2022.

\bibitem[Zhu et~al.(2023)Zhu, Yao, Sun, Li, and Liu]{zhu2023weak}
Zhaowei Zhu, Yuanshun Yao, Jiankai Sun, Hang Li, and Yang Liu.
\newblock Weak proxies are sufficient and preferable for fairness with missing sensitive attributes.
\newblock 2023.

\end{thebibliography}
